\documentclass[preprint,12pt]{elsarticle}

\usepackage{amsmath,amssymb,amsthm}
\usepackage{booktabs}
\usepackage{graphicx}
\usepackage{makecell}
\usepackage{listings}
\usepackage[hidelinks,pdfusetitle=false]{hyperref}
\hypersetup{pdftitle={A Contradiction Judge Counted Reported Teammate Errors as Hallucinations}}
\usepackage{xurl}

\journal{}
\biboptions{sort&compress}
\theoremstyle{plain}
\newtheorem{proposition}{Proposition}
\theoremstyle{definition}
\newtheorem{definition}{Definition}

\newcommand{\advWeaAsLlamaFB}{0.50}
\newcommand{\advWeaAsLlamaNS}{0.15}
\newcommand{\advWeaAsNS}{0.00}
\newcommand{\advWeaAtNS}{0.46}
\newcommand{\advWeaOrigNS}{0.96}

\newcommand{\asBootCI}{[\ensuremath{-}0.0425, \ensuremath{-}0.0008]}
\newcommand{\asCI}{[\ensuremath{-}0.0458, +0.0041]}

\newcommand{\asClean}{0.0050}
\newcommand{\asD}{\ensuremath{-}0.0208}

\newcommand{\asGemD}{\ensuremath{-}0.0063}

\newcommand{\asGptD}{\ensuremath{-}0.0192}

\newcommand{\asLlamaCI}{[\ensuremath{-}0.0446, +0.0488]}
\newcommand{\asLlamaD}{+0.0021}

\newcommand{\asMaxVsMeanCI}{[0.0019, 0.0218]}
\newcommand{\asMaxVsMeanD}{+0.0119}

\newcommand{\asP}{0.093}

\newcommand{\asQwenD}{\ensuremath{-}0.0646}

\newcommand{\asSsvpD}{\ensuremath{-}0.0052}

\newcommand{\asSwCI}{[\ensuremath{-}0.1033, +0.0033]}
\newcommand{\asSwD}{\ensuremath{-}0.0500}

\newcommand{\asSwSbD}{\ensuremath{-}0.0021}

\newcommand{\atCI}{[0.1531, 0.2877]}

\newcommand{\atD}{+0.2204}

\newcommand{\atLlamaCI}{[0.1399, 0.2403]}
\newcommand{\atLlamaD}{+0.1901}

\newcommand{\atSbCI}{[\ensuremath{-}0.0100, +0.0397]}
\newcommand{\atSbD}{+0.0149}

\newcommand{\atSsvpCI}{[0.0004, 0.0384]}
\newcommand{\atSsvpD}{+0.0194}

\newcommand{\attShareInjMax}{96}
\newcommand{\attShareInjMin}{39}

\newcommand{\bookAsFB}{0.80}

\newcommand{\bookAsNS}{0.96}
\newcommand{\bookAtFB}{0.20}
\newcommand{\cascadeBound}{0.208}
\newcommand{\dOrig}{5.95}

\newcommand{\decAsDestAir}{\ensuremath{-}0.0200}

\newcommand{\decMnDestAir}{+0.2042}

\newcommand{\decOrigDestAirPct}{77}

\newcommand{\eOneBCI}{[\ensuremath{-}0.0196, +0.0279]}
\newcommand{\eOneBD}{+0.0042}

\newcommand{\eThreeCI}{[\ensuremath{-}0.1267, +0.0112]}
\newcommand{\eThreeD}{\ensuremath{-}0.0578}

\newcommand{\eThreeLlamaCI}{[\ensuremath{-}0.1368, \ensuremath{-}0.0077]}
\newcommand{\eThreeLlamaD}{\ensuremath{-}0.0722}

\newcommand{\eThreeSwCI}{[\ensuremath{-}0.2755, +0.0088]}
\newcommand{\eThreeSwD}{\ensuremath{-}0.1333}

\newcommand{\eTwoCI}{[\ensuremath{-}0.0095, +0.0122]}
\newcommand{\eTwoD}{+0.0013}
\newcommand{\eTwoLlamaBootCI}{[0.0117, 0.0867]}
\newcommand{\eTwoLlamaCI}{[\ensuremath{-}0.0002, +0.0935]}
\newcommand{\eTwoLlamaD}{+0.0467}
\newcommand{\fleissAs}{0.763}
\newcommand{\fleissAsPrim}{0.737}
\newcommand{\fleissFour}{0.654}
\newcommand{\fleissFourPrim}{0.612}
\newcommand{\fleissMn}{0.775}
\newcommand{\fleissMnPrim}{0.792}

\newcommand{\mnCI}{[0.1522, 0.2470]}

\newcommand{\mnD}{+0.1996}

\newcommand{\mnGapToOrig}{0.065}

\newcommand{\mnLlamaCI}{[0.1377, 0.2466]}
\newcommand{\mnLlamaD}{+0.1922}

\newcommand{\mnSwCI}{[\ensuremath{-}0.0939, +0.0907]}
\newcommand{\mnSwD}{\ensuremath{-}0.0016}

\newcommand{\mnSwSbD}{+0.0125}

\newcommand{\nAttempts}{5{,}757}
\newcommand{\nConditions}{15}
\newcommand{\nFailedFB}{5}
\newcommand{\nFailedSb}{7}
\newcommand{\nFailedSsvp}{5}
\newcommand{\nFailedTotal}{17}

\newcommand{\nHaikuFB}{10}
\newcommand{\nHaikuNS}{25}
\newcommand{\nItems}{10{,}992}
\newcommand{\nItemsPrim}{1{,}680}
\newcommand{\nLlamaFB}{8}

\newcommand{\nLlamaTeamFB}{40}
\newcommand{\nLlamaTeamFBAssertDest}{2}
\newcommand{\nLlamaTeamNS}{100}
\newcommand{\nOutputs}{1{,}832}
\newcommand{\nReaderNonReport}{6}
\newcommand{\nReaderReports}{44}
\newcommand{\nRecords}{5{,}496}
\newcommand{\nRecovered}{5}
\newcommand{\nRetried}{17}
\newcommand{\nScenHaikuFB}{6}

\newcommand{\nTeamFB}{50}
\newcommand{\nTeamFBAssertAir}{0}
\newcommand{\nTeamFBAssertDest}{0}
\newcommand{\nTeamFBAttribDest}{49}
\newcommand{\nTeamFBCorrectDest}{1}

\newcommand{\nTies}{194}
\newcommand{\nTwoJudgeAdoptGem}{7}
\newcommand{\nTwoJudgeAdoptGpt}{6}
\newcommand{\origAsKappaMax}{0.414}
\newcommand{\origAsKappaMin}{0.355}

\newcommand{\origD}{+0.2650}

\newcommand{\origMaxErr}{\ensuremath{3.3\times10^{-5}}}

\newcommand{\origMaxVsMeanCI}{[\ensuremath{-}0.0450, +0.0066]}
\newcommand{\origMaxVsMeanD}{\ensuremath{-}0.0192}

\newcommand{\origMnAgree}{0.882}
\newcommand{\origMnKappa}{0.708}
\newcommand{\origMnKappaMax}{0.723}
\newcommand{\origMnKappaMin}{0.650}

\newcommand{\origSwCI}{[\ensuremath{-}0.0682, +0.1703]}
\newcommand{\origSwD}{+0.0510}

\newcommand{\origSwSbD}{\ensuremath{-}0.0688}

\newcommand{\pairAsMax}{0.785}
\newcommand{\pairAsMin}{0.748}

\newcommand{\pcBookPct}{97}

\newcommand{\pcCI}{[0.1115, 0.1285]}
\newcommand{\pcD}{+0.1200}

\newcommand{\pcSwCI}{[0.0064, 0.1103]}
\newcommand{\pcSwD}{+0.0583}

\newcommand{\planDurAsNS}{0.05}
\newcommand{\planDurAtNS}{0.92}
\newcommand{\planDurOrigNS}{1.00}
\newcommand{\pubCI}{[0.225, 0.305]}
\newcommand{\pubD}{+0.265}
\newcommand{\pubFB}{0.554}
\newcommand{\pubNS}{0.289}
\newcommand{\rawAsMax}{0.959}
\newcommand{\rawAsMin}{0.954}

\newcommand{\relFB}{\ensuremath{-}17}
\newcommand{\relFBCI}{[\ensuremath{-}34, 0]}
\newcommand{\scenBootCI}{[\ensuremath{-}0.0452, +0.0019]}
\newcommand{\scenFECI}{[\ensuremath{-}0.0355, +0.0159]}
\newcommand{\scenFEP}{0.40}
\newcommand{\scenFEd}{\ensuremath{-}0.0098}
\newcommand{\scenICC}{0.39}
\newcommand{\scenICCP}{0.035}
\newcommand{\scenLooHi}{\ensuremath{-}0.0130}
\newcommand{\scenLooLo}{\ensuremath{-}0.0276}
\newcommand{\scenNInBoth}{6}
\newcommand{\scenNInFB}{6}
\newcommand{\scenNInNS}{8}
\newcommand{\scenPermP}{0.39}
\newcommand{\shareHaiku}{\ensuremath{-}0.10}
\newcommand{\shareHaikuCI}{[\ensuremath{-}0.19, \ensuremath{-}0.004]}

\newcommand{\spendUSD}{2.71}
\newcommand{\teamAtDestAir}{+0.204}

\newcommand{\tieCI}{[\ensuremath{-}0.0450, +0.0108]}
\newcommand{\tieCIUpper}{+0.0108}
\newcommand{\tieD}{\ensuremath{-}0.0171}

\newcommand{\tostCIninety}{[\ensuremath{-}0.0412, \ensuremath{-}0.0005]}
\newcommand{\tostDmin}{0.0412}
\newcommand{\tostLlamaDmin}{0.0396}
\newcommand{\tostLlamaP}{0.007}
\newcommand{\tostMarginH}{0.1325}
\newcommand{\tostMarginQ}{0.0663}
\newcommand{\tostP}{0.001}
\newcommand{\tostSwDmin}{0.0935}
\newcommand{\tostSwP}{0.26}
\newcommand{\tostTieDmin}{0.0399}
\newcommand{\tostTieP}{0.001}

\newcommand{\unanCI}{[\ensuremath{-}0.0866, \ensuremath{-}0.0318]}
\newcommand{\unanD}{\ensuremath{-}0.0592}

\newcommand{\nvAisCI}{[\ensuremath{-}0.0480, +0.0155]}
\newcommand{\nvAisD}{\ensuremath{-}0.0163}
\newcommand{\nvAisKappa}{0.836}
\newcommand{\nvAisLlamaCI}{[+0.0031, +0.0844]}
\newcommand{\nvAisLlamaD}{+0.0437}
\newcommand{\nvAisLlamaMn}{+0.1380}
\newcommand{\nvAisLlamaMnCI}{[+0.0782, +0.1978]}
\newcommand{\nvAisMn}{+0.1442}
\newcommand{\nvAisMnCI}{[+0.1192, +0.1691]}
\newcommand{\nvAisSwCI}{[\ensuremath{-}0.0943, +0.0120]}
\newcommand{\nvAisSwD}{\ensuremath{-}0.0411}
\newcommand{\nvAisTostP}{0.003}
\newcommand{\nvFrFableCI}{[\ensuremath{-}0.1336, \ensuremath{-}0.0664]}
\newcommand{\nvFrFableD}{\ensuremath{-}0.1000}
\newcommand{\nvFrFableInvalid}{34}
\newcommand{\nvFrFableMn}{+0.2125}
\newcommand{\nvFrFableMnCI}{[+0.0782, +0.3468]}
\newcommand{\nvFrFableReqs}{80}
\newcommand{\nvFrFableTrunc}{8}
\newcommand{\nvFrOpusNeuCI}{[\ensuremath{-}0.1490, \ensuremath{-}0.0927]}
\newcommand{\nvFrOpusNeuD}{\ensuremath{-}0.1208}

\newcommand{\nvFrOpusNeuMn}{+0.1167}
\newcommand{\nvFrOpusNeuMnCI}{[+0.0689, +0.1645]}

\newcommand{\nvFrOpusOrigCI}{[\ensuremath{-}0.1419, \ensuremath{-}0.0747]}
\newcommand{\nvFrOpusOrigD}{\ensuremath{-}0.1083}
\newcommand{\nvFrOpusOrigFB}{0.0208}
\newcommand{\nvFrOpusOrigMn}{+0.2125}
\newcommand{\nvFrOpusOrigMnCI}{[+0.1316, +0.2934]}
\newcommand{\nvFrOpusOrigMnFB}{0.5583}
\newcommand{\nvFrOpusOrigMnNS}{0.3458}
\newcommand{\nvFrOpusOrigNS}{0.1292}
\newcommand{\nvFrReqs}{360}
\newcommand{\nvFrTrials}{5}
\newcommand{\nvFrTrunc}{316}
\newcommand{\nvHaikuAgentReqs}{2{,}880}
\newcommand{\nvHaikuAgentTrunc}{2{,}415}

\newcommand{\nvHaikuFlippFmt}{0.019}
\newcommand{\nvHaikuFlippLen}{0.013}
\newcommand{\nvHaikuInvTotal}{1}
\newcommand{\nvHaikuKappaPrior}{0.859}
\newcommand{\nvHaikuKappaVone}{0.870}
\newcommand{\nvHaikuKappaVthree}{0.906}
\newcommand{\nvHaikuKappaVtwo}{0.761}
\newcommand{\nvHaikuKappapFmt}{0.900}
\newcommand{\nvHaikuKappapLen}{0.931}
\newcommand{\nvHaikuVzeroCI}{[\ensuremath{-}0.0796, \ensuremath{-}0.0204]}
\newcommand{\nvHaikuVzeroD}{\ensuremath{-}0.0500}
\newcommand{\nvHaikuVzeroMn}{+0.1142}
\newcommand{\nvHaikuVzeroMnCI}{[+0.0708, +0.1575]}

\newcommand{\nvJudgeKappaVzero}{0.784}
\newcommand{\nvOneAgainst}{1}
\newcommand{\nvOneDecrease}{5}
\newcommand{\nvOneEquiv}{6}
\newcommand{\nvOneOutputs}{648}
\newcommand{\nvOneRequests}{7{,}776}
\newcommand{\nvPostSonOneCI}{[+0.0282, +0.0822]}
\newcommand{\nvPostSonOneD}{+0.0552}

\newcommand{\nvSonInvVthree}{442}
\newcommand{\nvSonInvVtwo}{296}

\newcommand{\nvSonOneCI}{[+0.0055, +0.0436]}
\newcommand{\nvSonOneD}{+0.0246}
\newcommand{\nvSonOneLlamaCI}{[+0.0040, +0.0689]}
\newcommand{\nvSonOneLlamaD}{+0.0365}
\newcommand{\nvSonReqs}{3{,}888}
\newcommand{\nvSonTruncReq}{977}
\newcommand{\nvSonVzeroCI}{[\ensuremath{-}0.0218, +0.0184]}
\newcommand{\nvSonVzeroD}{\ensuremath{-}0.0017}
\newcommand{\nvSonVzeroMn}{+0.1292}
\newcommand{\nvSonVzeroMnCI}{[+0.0929, +0.1654]}

\newcommand{\nvSonnetFlippFmt}{0.029}
\newcommand{\nvSonnetFlippLen}{0.025}
\newcommand{\nvSonnetKappaPrior}{0.830}
\newcommand{\nvSonnetKappaVone}{0.833}
\newcommand{\nvSonnetKappaVthree}{0.331}
\newcommand{\nvSonnetKappaVtwo}{0.685}
\newcommand{\nvSonnetKappapFmt}{0.863}
\newcommand{\nvSonnetKappapLen}{0.880}
\newcommand{\nvSpendTotal}{172.59}
\newcommand{\nvThreeBothNeuCI}{[\ensuremath{-}0.0971, \ensuremath{-}0.0675]}
\newcommand{\nvThreeBothNeuD}{\ensuremath{-}0.0823}

\newcommand{\nvThreeBothNeuMn}{+0.1161}
\newcommand{\nvThreeBothNeuMnCI}{[+0.0895, +0.1428]}
\newcommand{\nvThreeBothOrigCI}{[\ensuremath{-}0.0627, \ensuremath{-}0.0383]}
\newcommand{\nvThreeBothOrigD}{\ensuremath{-}0.0505}
\newcommand{\nvThreeBothOrigMeanFB}{0.0755}
\newcommand{\nvThreeBothOrigMeanNS}{0.1260}
\newcommand{\nvThreeBothOrigMn}{+0.0557}
\newcommand{\nvThreeBothOrigMnCI}{[+0.0321, +0.0794]}
\newcommand{\nvThreeBothOrigTostP}{0.0064}
\newcommand{\nvThreeDidAs}{\ensuremath{-}0.0318}
\newcommand{\nvThreeDidAsCI}{[\ensuremath{-}0.0504, \ensuremath{-}0.0131]}
\newcommand{\nvThreeDidMn}{+0.0604}
\newcommand{\nvThreeDidMnCI}{[+0.0256, +0.0952]}
\newcommand{\nvThreeEitherNeuCI}{[\ensuremath{-}0.0394, \ensuremath{-}0.0013]}
\newcommand{\nvThreeEitherNeuD}{\ensuremath{-}0.0203}
\newcommand{\nvThreeEitherNeuMn}{+0.2016}
\newcommand{\nvThreeEitherNeuMnCI}{[+0.1795, +0.2237]}
\newcommand{\nvThreeEitherOrigCI}{[+0.0024, +0.0299]}
\newcommand{\nvThreeEitherOrigD}{+0.0161}
\newcommand{\nvThreeEitherOrigMn}{+0.1542}
\newcommand{\nvThreeEitherOrigMnCI}{[+0.1357, +0.1726]}

\newcommand{\nvThreeHaikuOrigD}{\ensuremath{-}0.0385}

\newcommand{\nvThreeKappaJudges}{0.773}
\newcommand{\nvThreeMajCI}{[\ensuremath{-}0.0370, \ensuremath{-}0.0130]}
\newcommand{\nvThreeMajD}{\ensuremath{-}0.0250}

\newcommand{\nvThreeOpusCI}{[\ensuremath{-}0.0485, \ensuremath{-}0.0234]}
\newcommand{\nvThreeOpusD}{\ensuremath{-}0.0359}

\newcommand{\nvThreeSonInvalid}{666}
\newcommand{\nvThreeSonReqs}{1{,}280}
\newcommand{\nvThreeSonTrunc}{111}

\newcommand{\nvThreeSonnetOrigD}{+0.0042}

\newcommand{\nvThreeTrials}{40}
\newcommand{\nvWAgentReqs}{1{,}440}

\newcommand{\nvWFableCI}{[\ensuremath{-}0.1655, \ensuremath{-}0.1355]}
\newcommand{\nvWFableCut}{33}
\newcommand{\nvWFableD}{\ensuremath{-}0.1505}
\newcommand{\nvWFableFB}{0.0375}
\newcommand{\nvWFableInvalid}{45}
\newcommand{\nvWFableMn}{+0.2214}
\newcommand{\nvWFableMnCI}{[+0.1958, +0.2469]}
\newcommand{\nvWFableMnFB}{0.5844}
\newcommand{\nvWFableMnNS}{0.3630}
\newcommand{\nvWFableNS}{0.1880}

\newcommand{\nvWKappa}{0.768}
\newcommand{\nvWOpusCI}{[\ensuremath{-}0.1430, \ensuremath{-}0.1226]}
\newcommand{\nvWOpusCut}{13}
\newcommand{\nvWOpusD}{\ensuremath{-}0.1328}
\newcommand{\nvWOpusFB}{0.0167}
\newcommand{\nvWOpusInvalid}{2}
\newcommand{\nvWOpusMn}{+0.2307}
\newcommand{\nvWOpusMnCI}{[+0.2067, +0.2548]}
\newcommand{\nvWOpusMnFB}{0.5693}
\newcommand{\nvWOpusMnNS}{0.3385}
\newcommand{\nvWOpusNS}{0.1495}

\newcommand{\nvWReqsPerArm}{720}

\newcommand{\nvWTrials}{40}
\newcommand{\nvWTrunc}{305}
\newcommand{\nvWTruncFB}{304}
\newcommand{\nvWTruncNS}{1}
\newcommand{\nvWTruncPct}{21.2}
\newcommand{\nvXAgentReqs}{576}

\newcommand{\nvXFableCI}{[\ensuremath{-}0.1947, \ensuremath{-}0.1361]}
\newcommand{\nvXFableCut}{0}
\newcommand{\nvXFableD}{\ensuremath{-}0.1654}
\newcommand{\nvXFableFB}{0.0469}
\newcommand{\nvXFableInvalid}{0}
\newcommand{\nvXFableMn}{+0.1888}
\newcommand{\nvXFableMnCI}{[+0.1514, +0.2262]}
\newcommand{\nvXFableMnFB}{0.5768}
\newcommand{\nvXFableMnNS}{0.3880}
\newcommand{\nvXFableNS}{0.2122}

\newcommand{\nvXKappa}{0.755}
\newcommand{\nvXMaxFB}{6{,}576}
\newcommand{\nvXMaxNS}{3{,}528}
\newcommand{\nvXOpusCI}{[\ensuremath{-}0.1549, \ensuremath{-}0.1211]}
\newcommand{\nvXOpusCut}{1}
\newcommand{\nvXOpusD}{\ensuremath{-}0.1380}
\newcommand{\nvXOpusFB}{0.0182}
\newcommand{\nvXOpusInvalid}{0}
\newcommand{\nvXOpusMn}{+0.2279}
\newcommand{\nvXOpusMnCI}{[+0.1955, +0.2602]}
\newcommand{\nvXOpusMnFB}{0.5729}
\newcommand{\nvXOpusMnNS}{0.3451}
\newcommand{\nvXOpusNS}{0.1562}
\newcommand{\nvXOpusReqs}{256}
\newcommand{\nvXRefusals}{0}

\newcommand{\nvXStepMaxFBHi}{6{,}576}
\newcommand{\nvXStepMaxFBLo}{3{,}161}
\newcommand{\nvXStepMaxNSHi}{3{,}528}
\newcommand{\nvXStepMaxNSLo}{2{,}897}
\newcommand{\nvXTrials}{16}
\newcommand{\nvXTrunc}{0}
\newcommand{\nvYAFableCI}{[\ensuremath{-}0.1337, \ensuremath{-}0.0600]}

\newcommand{\nvYAFableD}{\ensuremath{-}0.0969}

\newcommand{\nvYAFableMn}{+0.2849}
\newcommand{\nvYAFableMnCI}{[+0.2211, +0.3487]}

\newcommand{\nvYAOpusCI}{[\ensuremath{-}0.1628, \ensuremath{-}0.1038]}

\newcommand{\nvYAOpusD}{\ensuremath{-}0.1333}

\newcommand{\nvYAOpusMn}{+0.2917}
\newcommand{\nvYAOpusMnCI}{[+0.2363, +0.3470]}

\newcommand{\nvYATrialsFB}{8}
\newcommand{\nvYATrialsNS}{5}
\newcommand{\nvYCutoff}{0}
\newcommand{\nvYDroppedRunFB}{0}
\newcommand{\nvYDroppedRunNS}{3}
\newcommand{\nvYEmptyRef}{3}
\newcommand{\nvYJudgeReqsFable}{104}
\newcommand{\nvYJudgeReqsOpus}{104}
\newcommand{\nvYPartialRef}{15}
\newcommand{\nvYPlanned}{8}
\newcommand{\nvYRFableCI}{[\ensuremath{-}0.1831, \ensuremath{-}0.0530]}

\newcommand{\nvYRFableD}{\ensuremath{-}0.1181}
\newcommand{\nvYRFableFB}{0.0278}

\newcommand{\nvYRFableMn}{+0.2153}
\newcommand{\nvYRFableMnCI}{[+0.1880, +0.2425]}

\newcommand{\nvYRFableNS}{0.1458}

\newcommand{\nvYROpusCI}{[\ensuremath{-}0.1976, \ensuremath{-}0.0941]}

\newcommand{\nvYROpusD}{\ensuremath{-}0.1458}
\newcommand{\nvYROpusFB}{0.0000}

\newcommand{\nvYROpusMn}{+0.2500}
\newcommand{\nvYROpusMnCI}{[+0.1428, +0.3572]}

\newcommand{\nvYROpusNS}{0.1458}
\newcommand{\nvYRTrialsFB}{3}
\newcommand{\nvYRTrialsNS}{3}

\newcommand{\nvYRefBookingFB}{9}
\newcommand{\nvYRefBookingNS}{4}
\newcommand{\nvYRefFB}{9}
\newcommand{\nvYRefNS}{9}

\newcommand{\nvYRefPctFB}{6.25}
\newcommand{\nvYRefPctNS}{6.82}

\newcommand{\nvYRefPlannerNS}{5}
\newcommand{\nvYRefStepFiveFB}{1}
\newcommand{\nvYRefStepFiveNS}{1}
\newcommand{\nvYRefStepFourFB}{2}
\newcommand{\nvYRefStepFourNS}{3}
\newcommand{\nvYRefStepThreeFB}{4}
\newcommand{\nvYRefStepThreeNS}{4}
\newcommand{\nvYRefStepTwoFB}{2}
\newcommand{\nvYRefStepTwoNS}{1}
\newcommand{\nvYRemainRegFB}{3}
\newcommand{\nvYRemainRegNS}{3}
\newcommand{\nvYRemainRunFB}{8}
\newcommand{\nvYRemainRunNS}{5}

\newcommand{\nvYSentFB}{144}
\newcommand{\nvYSentNS}{132}

\makeatletter\let\tabinput\@@input\makeatother
\newcommand{\earlier}{\cite{rodrigues2026drift}}

\makeatletter
\def\ps@pprintTitle{\let\@oddhead\@empty\let\@evenhead\@empty\def\@oddfoot{\hfill\thepage\hfill}\let\@evenfoot\@oddfoot}
\makeatother

\begin{document}

\begin{frontmatter}

\title{A Contradiction Judge Counted Reported Teammate Errors as Hallucinations: An Assertion-Aware Re-Score of One Multi-Agent LLM Harness}

\author[inst1]{Carson Rodrigues\corref{cor1}}
\ead{carson@celabe.com}
\cortext[cor1]{Corresponding author. ORCID: 0009-0001-7195-6742.}
\affiliation[inst1]{organization={Celabe},
            country={India}}

\begin{abstract}
LLM judges that score hallucination flag any statement that contradicts a reference fact. Agents in multi-agent LLM systems quote and dispute each other's claims, so such a judge can count a correct report of a teammate's error as a hallucination. An earlier version of this study, a preprint (arXiv:2606.21666v1), reported that broadcasting raised a hallucination rate from 0.492 to 0.658 (+34\%) and called it contamination; this version withdraws that claim, and the preprint's runs cannot be re-scored. An unpublished revision of that harness reported \pubNS{} to \pubFB{} (\pubD{}, 95\% CI \pubCI{}), and its saved outputs are re-scored here. Under an analysis plan fixed before any judge call, three judges from three vendors labeled each of six facts in \nOutputs{} outputs as asserted wrong, attributed to another source, asserted correct or absent. On values that outputs assert as their own, the broadcast effect was \asD{} (95\% CI \asCI{}) on \nHaikuFB{} Full-Broadcast trials: not distinguishable from zero, and equivalent to zero only within a post hoc margin of $\pm$\tostMarginQ{} and under the plan's prompt. On these LLM-judge labels, reports of a teammate's wrong value accounted for nearly all of the Haiku travel effect. Scored for mentions, the judges reproduced the effect (\mnD{}). With Llama agents, \nLlamaTeamFBAssertDest{} of \nLlamaTeamFB{} teammate outputs asserted the wrong destination. One of twelve Claude judge-prompt combinations gave an asserted increase (Sonnet 4.6, one paraphrase: \nvSonOneD{}, 95\% CI \nvSonOneCI{}). In \nvThreeTrials{} new trials per arm, judged by Claude models like the agents, the asserted difference was \nvThreeBothOrigD{} when both judges agreed and \nvThreeEitherOrigD{} (95\% CI \nvThreeEitherOrigCI{}) when either did. Main limit: the labels have no human validation; human labels for 120 outputs are pending. The outputs come from one harness and 13 hand-written scenarios, so the shares do not transfer.

\medskip

\par\noindent\rule{\linewidth}{0.4pt}\par\smallskip
{\small
\textbf{Correction note (version 2).} Version 1 (arXiv:2606.21666v1, 19 June 2026, ``Hallucination as Context Drift: Synchronization Protocols for Multi-Agent LLM Systems'') reported that broadcasting every agent's output to its teammates raised a hallucination rate from 0.492 (no synchronization) to 0.658 (+34\%, $p=0.0022$, $d=1.18$), which it called a contamination effect; that its protocol SSVP differed from no synchronization by a non-significant reduction (0.463, $-0.029$, $p=0.257$); and that SSVP scored significantly below broadcast (0.463 against 0.658) with 58\% fewer API calls. An independent re-run of version 1's design reproduced those rates; what is withdrawn is their interpretation. Version 1 defined its rate over agent assertions, but its judge may have counted statements inconsistent with ground truth without asking whether an output asserted them or reported them as a teammate's claim; its runs saved only the judge's summaries, so this cannot be checked and they cannot be re-scored. We re-scored the saved outputs of a later, unpublished revision of the study under an analysis plan fixed before any judge call. There the broadcast effect on the original judge's flags was \pubD{} (95\% CI \pubCI{}); the re-scoring judges, scoring mentions, reproduced it (\mnD{}), and the effect on values the outputs assert as their own was \asD{} (95\% CI \asCI{}, \nHaikuFB{} Full-Broadcast trials). No majority of three LLM judges labeled any of the \nTeamFB{} exposed teammate outputs (Haiku, Full-Broadcast, planner and advisor) as asserting the wrong destination (\nTeamFBAttribDest{} attributed it to another agent or source, \nTeamFBCorrectDest{} asserted the correct value); these labels have no human validation yet. The evidence is not uniform: with Llama agents \nLlamaTeamFBAssertDest{} of \nLlamaTeamFB{} teammate outputs asserted the wrong destination under Full-Broadcast, and in later Claude-judged trials the asserted difference was \nvThreeEitherOrigD{} (95\% CI \nvThreeEitherOrigCI{}) when either judge counted it (Section~\ref{sec:nov}).

\smallskip
\textbf{Withdrawn:} the contamination claim, that is, the reading of the broadcast increase as the spread of a wrong value between agents, which the re-scored outputs do not support (some adoption, with Llama agents and under some judges, is not excluded); and what rested on the broadcast score: that SSVP avoids that failure or beats broadcast significantly. \textbf{Contradicted:} version 1's case-study claim that broadcast agents converge silently on the wrong value; the re-score found that teammate outputs mostly reported it as another agent's claim. \textbf{Not withdrawn:} version 1's report that SSVP differed from no synchronization by a non-significant $-0.029$ ($p=0.257$); it is not evidence of a reduction. In the revision's re-scored outputs SSVP also stays near No-Sync on the asserted score (\asSsvpD{}, Table~\ref{tab:contrasts}). \textbf{Not re-tested:} the 58\% figure counts API calls, not judged outputs, so the re-score does not change it, but this version does not re-measure it and it no longer supports a claim that SSVP is better; the task-coherence drop and the non-replication in software that version 1 reported were explained by the withdrawn contamination reading and are not re-tested; the context divergence score and the rest of version 1 are likewise not re-tested and are not part of this version. Version 2 replaces the paper with this re-score study; later runs with Claude judges did not give a single answer (Section~\ref{sec:nov}).
\par}
\par\smallskip\noindent\rule{\linewidth}{0.4pt}
\end{abstract}

\begin{keyword}
LLM-as-a-judge \sep Hallucination evaluation \sep Measurement validity \sep Multi-agent systems \sep Large language models \sep Attribution
\end{keyword}

\end{frontmatter}

\section{Introduction}
\label{sec:intro}

Hallucination in the outputs of large language models (LLMs) is usually measured against reference facts: a detector or a judge model reads an output and flags statements that contradict a source or a known value \cite{maynez2020faithfulness,kryscinski2020factcc,laban2022summac,honovich2022true,min2023factscore}. LLM judges now do much of this scoring \cite{zheng2023judging,gu2024survey}, including in systems built from several cooperating agents \cite{wu2023autogen,hong2023metagpt,du2024debate}. There, surveys describe hallucination as a failure that can pass from agent to agent \cite{hallucsurvey2025}, benchmarks trace errors that propagate through the steps of agent workflows \cite{agenthallu2026}, and failure taxonomies list misalignment between agents \cite{cemri2025mast}.

Multi-agent outputs differ from single-model outputs in a way that matters for this kind of scoring. Agents read each other's messages, and their outputs quote, compare and dispute what teammates said. A planner that writes ``the booking record lists Lisbon, but the task says Barcelona'' mentions a wrong value without committing to it. A judge instructed to flag any statement incompatible with a fact flags that sentence as a contradiction of the destination. Such a judge measures how often a wrong value appears in an output. How often the output asserts it is a different quantity.

The two quantities separate most in comparisons whose conditions change how often agents see each other's claims, and communication protocols for agent teams differ in exactly that respect. A protocol that shows every agent its teammates' outputs will raise a mention-sensitive score whenever one teammate holds a wrong value, even if no agent adopts it, and the increase will read as error propagation.

This paper documents a case where that happened and measures the size of the distortion. An earlier version of this study, posted as a preprint \earlier{}, compared communication protocols in three-agent LLM teams with injected errors and reported that broadcasting every agent's full output to its teammates raised a hallucination rate. It read the increase as contamination: broadcast carries one agent's wrong value into its teammates' outputs. A later revision of the study, which was not published, rebuilt the experiment harness, added conditions and a second agent family, and scored outputs with a separate contradiction judge. It reported that broadcast raised a graded hallucination rate from \pubNS{} to \pubFB{} (difference \pubD{}, 95\% CI \pubCI{}). We call it the revision; its saved outputs are the ones re-scored here. A reading of the revision's broadcast teammate outputs suggested that most of them reported the wrong value as another agent's claim. Before re-scoring, we fixed an analysis plan with decision rules that state what the paper would claim under each outcome. Three judges from three vendors then labeled every saved output, fact by fact, as asserting a wrong value, attributing it to another source, asserting the correct value, or not addressing the fact.

The contributions are as follows.
\begin{enumerate}
\item \textbf{A measurement-validity result} (Sections~\ref{sec:res-null} and~\ref{sec:res-where}). On values an output asserts as its own, the broadcast effect is \asD{} (95\% CI \asCI{}), against \mnD{} \mnCI{} on the same judges' mention-sensitive score; the asserted effect is equivalent to zero within $\pm$\tostMarginQ{} (a quarter of the revision's effect) and stays so when scenario is modeled. None of the \nTeamFB{} exposed teammate outputs asserted the wrong destination, and \nTeamFBAttribDest{} attributed it to another agent or source. In these outputs, the revision's effect consists of such reports. Two limits come with it: the labels are LLM judgments with no human validation yet, and the equivalence rests on \nHaikuFB{} Full-Broadcast trials and a margin chosen after the result.
\item \textbf{An assertion-aware labeling protocol with checks a reader can rerun} (Sections~\ref{sec:protocol}, \ref{sec:res-agree} and~\ref{sec:res-control}): majority labels of three judges from three vendors, with Fleiss' $\kappa=\fleissAs$ on the asserted label over \nItems{} items; agreement with the original judge when the labels are scored for mentions ($\kappa=\origMnKappa$); a positive control in which the asserted score registers injected errors (\pcD{}, 95\% CI \pcCI{}); and the result under each judge alone.
\item \textbf{An accounting of the bias} (Section~\ref{sec:model}) that splits a mention-sensitive contrast into an asserted part and an attributed part and writes the attributed part as exposure times the probability of reporting. The attributed contrast was \atD{} for full broadcast and \atSsvpD{} and \atSbD{} for the two protocols that exchanged summaries instead of outputs (Section~\ref{sec:res-all}).
\item \textbf{Recommendations for scoring hallucination in multi-agent outputs} (Section~\ref{sec:discussion}).
\end{enumerate}
The contamination hypothesis motivated the re-score, and on the revision's outputs the re-score does not support it. The analysis plan named that outcome in advance (Section~\ref{sec:plan}), and we report it as such. All results come from one harness and 13 hand-written scenarios, and no human labels yet check the judges. The preprint (version 1 of arXiv:2606.21666) stated the contamination claim; this version replaces it.

\section{Related work}
\label{sec:related}

\subsection{Scoring hallucination and factual consistency}
Hallucination covers generated content that is unsupported by, or contradicts, a source or world knowledge \cite{maynez2020faithfulness,ji2023survey,huang2025survey}. Automatic scores check outputs against a source with entailment models \cite{kryscinski2020factcc,laban2022summac,honovich2022true}, decompose them into atomic claims that are checked against a knowledge source \cite{min2023factscore}, or test consistency across samples \cite{manakul2023selfcheckgpt}. Surveys of hallucination evaluation separate faithfulness to a source from factuality against world knowledge \cite{qi2024survey}, and a large comparison found that hallucination metrics often disagree with human judgments \cite{kulkarni2025mirage}. An output that reports another agent's wrong claim is faithful to that agent's message and contradicts the world, so a factuality judge flags it. All of these scores concern the content of a statement and do not ask whether the writer commits to it. In single-model generation the two seldom come apart. Attribution research asks a related question: whether a statement is supported by an identified source \cite{rashkin2023ais}, or whether the passages a model cites support the sentences that cite them \cite{gao2023alce}. The distinction in this paper concerns the speaker instead: whether the output asserts a value as its own or reports it as someone else's claim. In multi-agent outputs the two come apart often.

\subsection{Closest prior work}
The distinction between a speaker's commitment and a report of someone else's claim is studied in linguistics and in event-factuality annotation. FactBank marks, for each event, which source (the author or another) presents it as factual \cite{sauri2009factbank}. CommitmentBank rates how far a speaker is committed to the content of an embedded clause, and models of speaker commitment were evaluated on it \cite{jiang2019florence}. A recent study finds that LLMs do poorly at source-aware belief annotation on FactBank \cite{murzaku2025zeroshot}. These works model commitment in human-written text; none scores the hallucination of generated multi-agent output, and none compares a commitment-aware label with a contradiction judge on the same outputs. Benchmarks for agent hallucination, such as MIRAGE-Bench, judge an agent's next action against its context \cite{zhang2025miragebench}, and they do not separate an agent's own assertion from its report of a teammate's claim. A dated web search (30 October 2026; queries and results in the supplementary file \texttt{LITERATURE-SEARCH-2026-10-30.md}) found no earlier measurement of how often a contradiction judge counts reported conflicts as hallucinations in multi-agent outputs. It was a web search, not a systematic review of Scopus or Web of Science, so the absence is not proof. Our attributed label is a related construct to attribution in the style of AIS \cite{rashkin2023ais}; a claim-level judge of that kind was run on a subset of these outputs in November 2026 (Section~\ref{sec:nov}).

\subsection{LLM judges and their reliability}
LLMs are used as judges in place of human raters \cite{zheng2023judging,chiang2023alternative,gu2024survey}. Studies of their reliability report position bias \cite{wang2024fair}, a preference for their own outputs \cite{panickssery2024self}, and agreement with humans that varies widely across tasks and judges \cite{bavaresco2025judges,thakur2024judging}. Recent studies evaluate LLM judges of relevance on a new retrieval dataset \cite{fernandes2027normastcu} and propose ways to make LLM evaluators more reliable \cite{meng2027reliable}. These studies mostly concern judges of quality or preference. The failure reported here is a mismatch between the question a factuality judge is asked and the construct a study claims to measure. It does not depend on the quality of the judge: the labels of three capable judges reproduce it when they are scored the way the original judge scored (Section~\ref{sec:res-agree}).

\subsection{Measurement validity in AI evaluation}
Measurement frameworks from the social sciences separate a construct from its operationalization and ask whether the operationalization measures the construct \cite{jacobs2021measurement,wallach2025position}. Xiao et al.~\cite{xiao2023measurement} applied this view to metrics for natural language generation, and Bowman and Dahl~\cite{bowman2021fix} examined the validity of benchmarks in natural language understanding. The case reported here is a construct-validity failure in which the gap between the construct (asserted falsehood) and the operationalization (any incompatible statement) is correlated with the treatment. The gap therefore biases a comparison instead of adding noise to it.

\subsection{Multi-agent LLM systems and error propagation}
Frameworks coordinate role-specialized agents through conversation \cite{wu2023autogen,hong2023metagpt}, and multi-agent debate exchanges answers to improve factuality \cite{du2024debate}. A failure taxonomy of multi-agent systems lists inter-agent misalignment as a distinct category \cite{cemri2025mast}, studies of faulty agents measure how one bad agent degrades a team \cite{huang2024resilience}, a survey of agent hallucination describes errors that pass between agents through their messages \cite{hallucsurvey2025}, and a benchmark locates the step at which an error enters an agent's trajectory \cite{agenthallu2026}. Any study that scores intermediate agent outputs against reference facts faces the measurement problem described here once those outputs discuss each other.

\section{The study being re-scored}
\label{sec:setting}

\subsection{Task, agents and injected errors}
Three agents produce a five- or seven-day travel plan for two travelers: a planner (day-by-day itinerary), a booking agent (flights, hotel, transport) and an advisor (weather, dining, local conditions). Each trial draws one of eight scenarios with its own destination, dates, budget, trip length and an alternative destination. The six ground-truth facts are the destination, the arrival airport, the dates, the budget, the weather and the trip length. Three errors are injected. The booking agent's system prompt and structured context name the alternative destination and its airport code and instruct it to report these details in all updates. The advisor's system prompt carries weather for the wrong month. The planner's structured context lists only part of the trip, while its system prompt states the correct length. A second task, software project planning with five scenarios, follows the same design with a project manager, a developer and a QA engineer, and the developer holds a wrong platform and technology stack in its system prompt. A clean condition injects no errors, although the clean booking and developer agents still name the alternative airport and platform in their history and goals (Section~\ref{sec:limits}).

\subsection{Communication conditions}
Each trial runs six steps, and at each step every agent writes one output of at most 512 tokens. Under \emph{No-Sync} the agents exchange no messages. Under \emph{Full-Broadcast} each agent's output is delivered to the other two (the last four received messages enter the next prompt), and every pair also runs a reconciliation step. Under \emph{Selective-Broadcast} every pair runs reconciliation at steps 2 and 4 and outputs are not delivered. \emph{SSVP} runs the same reconciliation only for pairs whose embedding-based divergence score exceeds a threshold, and \emph{SSVP-max} uses the largest pairwise score, rather than the mean, to decide. In reconciliation each agent of a pair writes a summary, reads its partner's summary with an instruction to identify contradicting beliefs, and writes an adjudication whose first 100 characters enter its history. The agents ran on Claude Haiku 4.5 (OpenRouter slug \texttt{anthropic/claude-haiku-4-5}) and Llama-3.3-70B (\texttt{meta-llama/llama-3.3-70b-instruct}) through OpenRouter at temperature 0.7, with one seed per condition. The saved trial files record \texttt{agent\_model} as empty, because the client fixes the slug; the slugs above come from the repository's README and commit messages, and OpenRouter slugs are aliases that do not record a model snapshot. The agent runs took place on 24 and 25 July 2026 (the Haiku and Llama travel folders were committed on those two days and the software folders on 25 July); the file system keeps no finer run times. We did not record the cost of the agent runs per condition. No instruction, fact or score from the evaluation reached an agent: a scan of the 312 prompt parts of every agent in every scenario, with and without injected errors, found none of the judges' fact sentences and none of the words ``hallucination'', ``judge'', ``rubric'' or ``ground truth'' (\texttt{leakage\_check\_2026\_10\_30.py}), and the judge ran after the sixth step. The prompts do ask agents to flag conflicts (the task prompt tells them to ``flag the discrepancy explicitly''), which invites the reporting that the re-score measures (Section~\ref{sec:limits}). The supplementary run log (\texttt{RUN-LOG.md}) lists the 41 result folders, of which 15 enter this paper; the others are smoke tests, pilots and earlier-pipeline runs, none dropped for its result. Trial numbers were set by the run commands and by OpenRouter credit, not by an interim analysis.

\subsection{Original metric and result}
\label{sec:original}
For each trial, eight outputs are judged: three planner, three booking and two advisor outputs from steps 2 to 4 (project manager, developer and QA engineer in software). The original judge, \texttt{gpt-4o-mini} at temperature 0.7, received the six facts and one output and returned the facts the output contradicts, with the instruction that any incompatible statement counts. The per-output score is the number of flagged facts divided by six, the graded hallucination rate of a trial is the mean over its eight outputs, and the unit of analysis is the trial. With Haiku agents in the travel task, Full-Broadcast raised this rate from \pubNS{} (\nHaikuNS{} trials) to \pubFB{} (\nHaikuFB{} trials), a difference of \pubD{} (95\% CI \pubCI{}; Cohen's $d=\dOrig$ over trial-level rates), and the destination and airport facts accounted for \decOrigDestAirPct\% of the increase. The revision reported this as contamination, together with the same increase with Llama agents. The preprint \earlier{} had made the same claim from earlier runs, which used a previous version of the harness. The preprint defined its rate over agent assertions and described its scoring as structured fact checks for 78\% of evaluated assertions and an LLM judge for the rest, while our record of the judge it used is a binary per-output contradiction judge built on the agents' own model; we cannot check which description is right, and either way the judge may have counted reports as flagged statements. Those runs saved the judge's summary of each output rather than the output, so they cannot be re-scored, and their estimate is open to the same bias; we do not claim that it was biased. Of the Haiku travel runs re-scored here, \nFailedTotal{} trials (\nFailedFB{} Full-Broadcast, \nFailedSsvp{} SSVP and \nFailedSb{} Selective-Broadcast) stopped with an API error when the account's credit ran out; they left no outputs and were excluded from the revision's analysis as well as from this one.

\subsection{What prompted the re-score}
A reading of the \nTeamFB{} Haiku Full-Broadcast teammate outputs (planner and advisor) classed \nReaderReports{} as reporting the wrong destination as the booking agent's claim and \nReaderNonReport{} as going further, and two LLM judges asked a binary question (adopts or reports) classed \nTwoJudgeAdoptGpt{} and \nTwoJudgeAdoptGem{} of the \nTeamFB{} as adopting. The reading was prepared with an LLM tool and confirmed by the author, so it is not an independent human reading, and it covered one fact in one cell. Two judge prompts were written for these outputs besides the original study's contradiction prompt: the binary adopt-or-report prompt of that two-judge check, run once per judge on the 50 outputs, and the assertion-aware prompt of Section~\ref{sec:protocol}, fixed in the plan before any call and not changed afterward. The repository history holds one version of each; drafts that were never committed are not recorded, so the count is of recorded versions. Six further prompt variants were registered and run in November 2026 (Section~\ref{sec:nov}). It showed that the headline could include reports, not how much of it did.

\section{Mention and assertion: an accounting of the bias}
\label{sec:model}

The decision rules of Section~\ref{sec:protocol} refer to two scores. This section defines them and shows how they separate in a comparison between protocols.

\begin{definition}[Labels and rates]
For an output $o$ and a fact $k$ with ground-truth value $x^*_k$, the label $\ell(o,k)$ is one of $\mathrm{AW}$ (the output asserts a value incompatible with $x^*_k$ as its own), $\mathrm{AT}$ (an incompatible value appears only as what another agent, record or source says, or as one side of an open conflict), $\mathrm{AC}$ (every value given is compatible with $x^*_k$) and $\mathrm{AB}$ (the fact is not addressed). Over the $m$ facts and the judged outputs $O$ of a trial, the asserted rate is $H_A=\frac{1}{m|O|}\sum_{o\in O}\sum_{k}\mathbf{1}[\ell(o,k)=\mathrm{AW}]$, the attributed rate $H_T$ is defined in the same way with $\mathrm{AT}$, and the mention-sensitive rate is $H_M=H_A+H_T$.
\end{definition}

A contradiction judge that flags any incompatible statement estimates $H_M$. If the construct of interest is asserted falsehood, $H_T$ is measurement error. For two protocols $P$ and $Q$,
\begin{equation}
\Delta H_M = \Delta H_A + \Delta H_T ,
\label{eq:split}
\end{equation}
so a mention-sensitive contrast estimates the asserted contrast with bias $\Delta H_T$. The bias vanishes only if the two protocols produce the same rate of outputs that report wrong values without asserting them. Protocols differ precisely in how much of each other's work agents see, so this condition is unlikely to hold.

The following statement makes the dependence on exposure explicit.

\begin{proposition}[Exposure term]
\label{prop:exposure}
Let agent $r$ hold a wrong value $v$ of fact $k$ that is fixed by its instructions, let $C(k)$ contain $k$ and the facts whose correct value depends on $k$, and suppose that under $Q$ no other agent's output mentions a wrong value of any fact in $C(k)$. Under $P$, let agent $i\neq r$ be exposed to $v$ with probability $e_i$, and let an exposed output of $i$ assert a wrong value of $l\in C(k)$ with probability $\alpha_{il}$ and only attribute one with probability $\beta_{il}$. Let $w_i$ be agent $i$'s share of the judged outputs. Then the teammates' contribution to the contrast on $C(k)$ is
\begin{equation}
\Delta H_A=\frac{1}{m}\sum_{l\in C(k)}\sum_{i\neq r}w_i e_i\alpha_{il},\qquad
\Delta H_T=\frac{1}{m}\sum_{l\in C(k)}\sum_{i\neq r}w_i e_i\beta_{il},
\label{eq:exposure}
\end{equation}
and their sum, the teammates' contribution to $\Delta H_M$, satisfies
\begin{equation}
\Delta H_A+\Delta H_T\le \frac{|C(k)|\,(1-w_r)}{m}.
\label{eq:bound}
\end{equation}
\end{proposition}
\begin{proof}
Under $Q$ the teammates' labels on $C(k)$ are neither $\mathrm{AW}$ nor $\mathrm{AT}$ by assumption. Under $P$ an output of $i$ carries $\mathrm{AW}$ on $l$ with probability $e_i\alpha_{il}$ and $\mathrm{AT}$ with probability $e_i\beta_{il}$. Averaging over outputs with weights $w_i$ and over the $m$ facts gives Eq.~(\ref{eq:exposure}). Because $\alpha_{il}+\beta_{il}\le1$, $e_i\le1$ and $\sum_{i\neq r}w_i=1-w_r$, the sum is at most $|C(k)|(1-w_r)/m$.
\end{proof}

Three consequences follow. First, a mention-sensitive contrast depends on $\alpha+\beta$ and cannot tell adoption from reporting: with $\alpha=0$ it can reach its upper bound while no teammate asserts anything. Second, the bias scales with the change in exposure, so protocols that pass only summaries between agents, or nothing, should carry little of it. Third, the source agent $r$ may switch between asserting and reporting its own wrong value once it sees teammates disagree. Such a switch lowers $H_A$ and raises $H_T$ by the same amount and leaves $H_M$ unchanged. The proposition is an accounting identity written after the original data were collected; it does not predict $\alpha$ or $\beta$, which Section~\ref{sec:results} measures. In the travel task, $w=(3/8,3/8,2/8)$ for the planner, the booking agent and the advisor, $r$ is the booking agent, and $C(\text{destination})$ contains the destination and the airport, so the bound is $2\times(5/8)/6=\cascadeBound$.

\section{Re-score protocol}
\label{sec:protocol}

\subsection{Analysis plan and decision rules}
\label{sec:plan}
The analysis plan (\texttt{ASSERTION\_RESCORE\_PROTOCOL.md} in the supplementary material) was committed at 22:49:24 UTC+05:30 on 12 October 2026, and the first judge call followed nine seconds later. It fixed the outputs, judges, prompt, labels, aggregation, metrics, contrasts and the decision rules below. It was committed to a private repository, so it is an analysis plan fixed before the calls and not a public preregistration \cite{nosek2018preregistration}. Let $D$ be the asserted-only difference between Full-Broadcast and No-Sync for Haiku agents in the travel task, $[L,U]$ its Welch 95\% interval, and $D_0=\origD$ the revision's difference recomputed from the stored flags.
\begin{itemize}
\item R0: if Fleiss' $\kappa$ for the binary asserted label over all items is below 0.40, the judges cannot measure the construct, and no other rule applies.
\item R1: if $L>0$ and $D\ge0.75\,D_0$, the revision's headline stands.
\item R2: if $L>0$ and $D<0.75\,D_0$, the paper is rewritten to lead with $D$.
\item R3: if $L\le0$, the effect of broadcast on asserted wrong values is not distinguishable from zero, and the paper is reframed as a measurement-validity result.
\end{itemize}
The rule is computed for the primary indicator and for a tie sensitivity, and the more conservative outcome applies. Secondary rules fixed the wording of the sentence on teammate adoption (S1) and of the Llama replication (S2), the reference point to use if the new judges' mention-sensitive contrast differs from $D_0$ by more than 0.08 (S3), and the condition under which the human subset may be called a check of the judges (S4). The plan also recorded our expectation, from a recount of the original flags, that $D$ would be about $+0.07$ to $+0.08$, which would have meant R2. The expectation was wrong.

\subsection{Outputs and judges}
Every output that the original judge scored in the Haiku and Llama runs was re-scored: \nOutputs{} outputs in \nConditions{} conditions (Table~\ref{tab:conditions}). The Qwen runs of the revision were excluded, because a local judge of the same model family had scored them and traces exist for only two of their four broadcast trials. The three judges come from three vendors, none of which made an agent model being judged: \texttt{gpt-4.1-mini} (OpenAI, also the vendor of the original judge), \texttt{gemini-2.5-flash} (Google) and \texttt{qwen3-235b-a22b-2507} (Alibaba, the non-thinking variant). Each was called once per output through OpenRouter at temperature 0, with a fixed seed, a 400-token limit and JSON output, and up to three attempts in all when a call failed or returned unparseable output; the provider that served each call is stored with its response. A judge saw the team description, the agent's role, the six facts as the original judge received them, and the output. It did not see the condition, the model, the trial or any earlier label. The calls ran on 12 October 2026 between 22:49 and 23:37 (UTC+05:30).

\subsection{Prompt and labels}
The prompt (\ref{app:prompt}) defines the four labels of Section~\ref{sec:model} and asks for one label per fact, with a precedence order ($\mathrm{AW}$, then $\mathrm{AT}$, then $\mathrm{AC}$, then $\mathrm{AB}$) for outputs to which more than one could apply, and a short verbatim quote as evidence for every $\mathrm{AW}$ or $\mathrm{AT}$ label. An output that plans on a wrong value anywhere is $\mathrm{AW}$ even if it gives the correct value elsewhere, and an output that gives a wrong value only as another source's claim is $\mathrm{AT}$ even if it also prefers the correct value.

\subsection{Aggregation, metrics and statistics}
An item, that is an (output, fact) pair, carries the majority label if at least two judges give that label; otherwise it is a tie (\nTies{} of \nItems{} items). The primary indicator is a majority $\mathrm{AW}$, with ties counted as not asserted; a tie sensitivity counts as asserted every tie that contains an $\mathrm{AW}$ vote. The mention-sensitive indicator is a majority answering $\mathrm{AW}$ or $\mathrm{AT}$, which is the question the original judge was asked, and the attributed-only indicator is mention-sensitive and not asserted. Each indicator gives a per-output score (labeled facts divided by six) and a trial score (the mean over eight outputs), as in the original metric. Contrasts are differences of trial means with Welch's $t$ interval, a percentile bootstrap interval over trials (20,000 resamples) and a Mann--Whitney test. All 15 contrasts of the original analysis are recomputed for every indicator, for each judge alone, and for variants that require one judge or all three. Intervals are not adjusted for multiple comparisons: the decision rules use the one primary contrast, and the other contrasts and variants are reported in full as descriptions. Agreement is measured with Fleiss' $\kappa$ \cite{fleiss1971} over the items with three valid labels and with pairwise Cohen's $\kappa$ \cite{cohen1960}. The analysis code reproduces every trial's stored original rate from the stored per-output flags to within \origMaxErr{}.

\subsection{Checks added after the result}
\label{sec:added}
Two analyses were added on 30 October 2026, after rule R3 had been applied, and are not in the plan. First, because the headline is a null, we test equivalence to zero with two one-sided Welch tests (TOST) against a margin of $\delta=0.25\,D_0=\tostMarginQ$, a quarter of the revision's effect; $0.5\,D_0=\tostMarginH$ and an absolute 0.05 are reported beside it, as is the smallest margin at which equivalence holds. The margin comes from $D_0$, not from the new estimate, but it was chosen after the result was known. Second, each trial draws one of the scenarios, so trials of one scenario are not independent and the arms are unbalanced on scenario. We re-estimate with scenario fixed effects and a cluster-robust interval over scenarios ($t$ with $G-1$ degrees of freedom), a permutation test that shuffles the arm label within scenarios (10,000 draws), a two-stage cluster bootstrap (scenarios, then trials; 10,000 draws) and leave-one-scenario-out. The code is \texttt{rigor\_fixes\_2026\_10\_30.py}; the scoring and aggregation code has 25 unit tests with hand-checked cases (\texttt{test\_assertion\_rescore\_units.py}).

\subsection{Human subset}
\label{sec:human}
Before any judge call, the plan drew 120 of the outputs for human labeling, stratified by condition and role (30 Haiku Full-Broadcast, 30 Haiku No-Sync, 15 and 15 Llama, 10 and 10 software, 5 SSVP and 5 Selective-Broadcast), and fixed the labeling sheet, the instructions and the key. The annotator may not be the author and may not use a model. Under rule S4, if Cohen's $\kappa$ between the annotator and the judge majority on the binary asserted label (720 items) is at least 0.60, the judge score may be described as checked against a human annotator; otherwise the annotator's own rates are reported next to the judges'. If the annotator's rate of asserted destinations on the Haiku Full-Broadcast teammate outputs differs from the judges' rate by more than 0.15, the decision of Section~\ref{sec:plan} is re-examined with the annotator's labels. These labels have not yet been collected (Section~\ref{sec:limits}).

\subsection{Deviations}
Two changes were made after the first call. D1: \nRecovered{} of the \nRecords{} judge records hold replies that were cut off after all six labels, and the analysis reads those labels by pattern. D2: \nRetried{} Qwen records with three failed provider attempts were called again once, after the analysis as originally written had been saved, and all of them returned labels. The primary result is identical with and without each deviation. The re-score made \nAttempts{} calls and cost US\$\spendUSD{}.

\section{Results}
\label{sec:results}

\subsection{The judges agree on what an output asserts}
\label{sec:res-agree}
Over all \nItems{} items, Fleiss' $\kappa$ was \fleissAs{} for the binary asserted label, \fleissMn{} for the binary mention label and \fleissFour{} for the four labels. On the \nItemsPrim{} items of the primary cells (Haiku travel, No-Sync and Full-Broadcast) the values were \fleissAsPrim{}, \fleissMnPrim{} and \fleissFourPrim{}. Pairwise $\kappa$ for the asserted label ranged from \pairAsMin{} to \pairAsMax{} (raw agreement \rawAsMin{} to \rawAsMax{}). These values indicate substantial agreement on the usual reading of $\kappa$ \cite{landis1977}, and rule R0 does not apply. Agreement among LLM judges shows reliability, not validity; the human subset (Section~\ref{sec:human}) is the planned check of validity, and its labels are still pending.

Against the original judge's per-fact flags, the majority mention label agreed with $\kappa=\origMnKappa$ (raw agreement \origMnAgree{}). Per judge, the mention label gave $\kappa$ from \origMnKappaMin{} to \origMnKappaMax{} and the asserted label from \origAsKappaMin{} to \origAsKappaMax{}. The new judges reproduce the original judge when they are scored the way it scored, and they depart from it where the labels separate assertion from attribution. The reproduction is not exact: the mention-sensitive rate lies below the original judge's rate in every condition with injected errors, with the largest gaps in the software task (Table~\ref{tab:conditions}), because the original judge also flagged items that the new majority labels correct, absent or tied.

\subsection{Broadcast does not raise the rate of asserted wrong values}
\label{sec:res-null}

\begin{table}[t]
\caption{Trial-mean graded rates by condition. Asserted, attributed-only and mention-sensitive use the majority of three judges (asserted plus attributed-only equals mention-sensitive); Original is the original judge's rate reconstructed from its stored flags. The last column is the attributed share of the mention-sensitive rate, a descriptive ratio that was not in the plan.}
\label{tab:conditions}
\centering
\footnotesize
\setlength{\tabcolsep}{4pt}
\begin{tabular}{lcccccc}
\toprule
Condition & $n$ & Asserted & Attributed & Mention & Original & Attr.\ share \\
\midrule
\tabinput tab_conditions.tex 
\bottomrule
\end{tabular}
\end{table}

\begin{table}[t]
\caption{Full-Broadcast minus No-Sync, Haiku agents, travel task (\nHaikuNS{} and \nHaikuFB{} trials), under every indicator. Welch and bootstrap columns are 95\% intervals; $p$ is from Welch's test. Asserted, attributed-only and mention indicators are majorities of the three judges unless a judge is named; the first row is the primary indicator on which the decision rule was computed.}
\label{tab:variants}
\centering
\scriptsize
\setlength{\tabcolsep}{4pt}
\begin{tabular}{lcccc}
\toprule
Indicator & Difference & Welch CI & Bootstrap CI & $p$ \\
\midrule
\tabinput tab_variants.tex 
\bottomrule
\end{tabular}
\end{table}

On the primary indicator, Full-Broadcast minus No-Sync was \asD{} (Welch 95\% CI \asCI{}, bootstrap \asBootCI{}, $p=\asP$; Table~\ref{tab:variants}). The Welch lower bound is below zero under the primary indicator and under the tie sensitivity (\tieD{}, \tieCI{}), so rule R3 applies. The bootstrap interval lies just below zero and the Welch interval includes zero, so we read the result as no increase; a decrease is not established under the primary indicator, although two variants give one whose interval excludes zero (requiring all three judges to agree, \unanD{} \unanCI{}, and the Qwen judge alone, \asQwenD{}). Under both indicators the Welch interval excludes an asserted increase larger than \tieCIUpper{}, against the revision's \pubD{}. No variant of this analysis gives a positive asserted increase whose interval excludes zero. Each judge alone gives \asGptD{}, \asGemD{} and \asQwenD{}, and requiring all three judges to agree gives \unanD{} \unanCI{}.

The null is also an equivalence, with two caveats. It rests on only \nHaikuFB{} Haiku Full-Broadcast trials, and the margin was chosen after the result was known (Section~\ref{sec:added}); we call it a post hoc margin throughout. Against the margin of \tostMarginQ{}, the 90\% Welch interval \tostCIninety{} lies inside the margin ($p_{\mathrm{TOST}}=\tostP$), and equivalence holds down to a margin of \tostDmin{}, which is below the absolute 0.05 that does not depend on $D_0$; with the tie sensitivity the values are $p=\tostTieP$ and \tostTieDmin{}. In relative terms the asserted rate under Full-Broadcast differs from the No-Sync rate by \relFB\% (bootstrap 95\% CI \relFBCI{}\%). The \nHaikuFB{} Full-Broadcast trials cover \scenNInFB{} scenarios and the No-Sync trials \scenNInNS{}, and scenario matters: the intraclass correlation of the No-Sync trial scores by scenario is \scenICC{} ($p=\scenICCP$). With scenario fixed effects the contrast is \scenFEd{} (cluster-robust 95\% CI \scenFECI{} over \scenNInBoth{} scenarios seen in both arms, $p=\scenFEP$), the within-scenario permutation test gives $p=\scenPermP$, the cluster bootstrap gives \scenBootCI{}, and leaving out any one scenario gives \scenLooLo{} to \scenLooHi{}. The conclusion does not depend on the unit of analysis. The new trials of Section~\ref{sec:nov} do not reproduce this null as one result: with the original prompts the asserted difference was \nvThreeBothOrigD{} when both Claude judges agree and \nvThreeEitherOrigD{} when either does, so the null is reported here for the earlier trials and judges only. Equivalence is established for Haiku and Llama travel (Llama: $p_{\mathrm{TOST}}=\tostLlamaP$, smallest margin \tostLlamaDmin{}) and not for software, where the cells are small (smallest margin \tostSwDmin{}, $p=\tostSwP$ at \tostMarginQ{}); Table~\ref{tab:cluster} in \ref{app:cluster} gives all three.

On the same judges' mention-sensitive score the contrast was \mnD{} \mnCI{}, which is \mnGapToOrig{} below the revision's value and within the limit of rule S3, and the attributed-only contrast was \atD{} \atCI{}. The asserted share of the mention-sensitive increase was \shareHaiku{} (bootstrap 95\% CI \shareHaikuCI{}). With Llama agents the pattern was the same: \asLlamaD{} \asLlamaCI{} on the asserted score and \mnLlamaD{} \mnLlamaCI{} on the mention-sensitive score. Under rule S2, the revision's claim that the increase recurred with Llama agents holds only for the mention-sensitive score.

\subsection{Where the mention-sensitive increase comes from}
\label{sec:res-where}

\begin{figure}[t]
\centering
\includegraphics{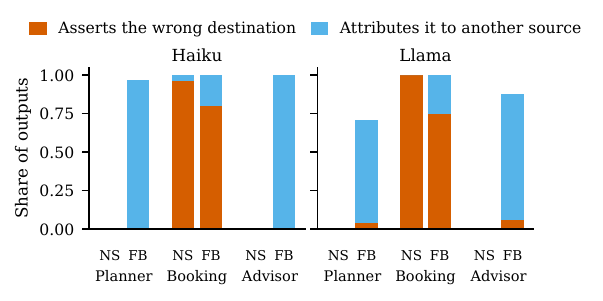}
\caption{Destination fact, travel task: share of each agent's judged outputs whose majority label asserts the wrong destination (dark) or attributes it to another source (light), under No-Sync (NS) and Full-Broadcast (FB). Outputs per agent (planner, booking, advisor): Haiku NS 75, 75, 50 and FB 30, 30, 20; Llama NS 60, 60, 40 and FB 24, 24, 16. Under broadcast the planner and advisor mention the wrong destination in most outputs and assert it in none (Haiku) or few (Llama).}
\label{fig:destination}
\end{figure}

On the destination and airport facts the mention-sensitive contrast was \decMnDestAir{} and the asserted contrast \decAsDestAir{}. No Haiku Full-Broadcast teammate output asserted the wrong destination or the wrong airport (\nTeamFBAssertDest{} and \nTeamFBAssertAir{} of \nTeamFB{}); the majority labeled the destination $\mathrm{AT}$ in \nTeamFBAttribDest{} of them and $\mathrm{AC}$ in \nTeamFBCorrectDest{} (Fig.~\ref{fig:destination}). The teammates' attributed contribution to the destination and airport components was \teamAtDestAir{}, against the bound of \cascadeBound{} from Proposition~\ref{prop:exposure}: broadcast came close to the largest mention-sensitive increment possible on these two facts with no adoption at all. Under rule S1 the sentence that teammates mostly reported the value rather than adopted it stands, and the judges support a stronger one: no majority of the three judges labeled any of them as asserting it, which is a statement about LLM labels and not a human-verified finding that none adopted it. With Llama agents, \nLlamaTeamFBAssertDest{} of \nLlamaTeamFB{} teammate outputs asserted the wrong destination under Full-Broadcast, against none of \nLlamaTeamNS{} under No-Sync.

The asserted wrong values came from the agents whose instructions held them. The booking agent asserted the wrong destination in \bookAsNS{} of its Haiku No-Sync outputs and \bookAsFB{} of its Full-Broadcast outputs, and in the remaining \bookAtFB{} it reported the conflict instead, which is the third consequence of Proposition~\ref{prop:exposure}. In checks that were not in the plan, restricting the asserted score to the booking agent gave a broadcast contrast of \eThreeD{} \eThreeCI{}, and restricting it to the teammates gave \eTwoD{} \eTwoCI{}. With Llama agents the same checks gave \eThreeLlamaD{} \eThreeLlamaCI{} for the booking agent and \eTwoLlamaD{} \eTwoLlamaCI{} for the teammates (bootstrap \eTwoLlamaBootCI{}). The Llama teammate increase includes the two destination adoptions and the advisor asserting its own wrong-month weather more often under broadcast (in \advWeaAsLlamaNS{} of its No-Sync outputs and \advWeaAsLlamaFB{} of its Full-Broadcast outputs), so with Llama agents broadcast may have raised asserted errors among the teammates slightly while lowering them in the booking agent. In software, the developer alone gave \eThreeSwD{} \eThreeSwCI{}, and no project manager or QA output under No-Sync or Full-Broadcast asserted a wrong value of any fact.

The original judge also counted reports on facts that broadcast did not change. Under No-Sync, the original judge flagged the planner's trip length in \planDurOrigNS{} of its outputs, but the planner asserted a wrong length in only \planDurAsNS{} and attributed one in \planDurAtNS{}, because it reported the gap between its partial itinerary and the stated trip length. The advisor's weather fact was flagged in \advWeaOrigNS{} of its outputs and asserted wrong in \advWeaAsNS{} (attributed in \advWeaAtNS{}).

Every output that the earlier reading did not class as a report (\nReaderNonReport{} of \nTeamFB{}) carries the majority label $\mathrm{AT}$. Three of them attribute the wrong city to the wrong agent, which is still a report. Three are borderline, for example a planner's own summary line ``Destination: Lisbon, Portugal (per Booking Agent)''. Counting all \nReaderNonReport{} as asserting both the destination and the airport gives \eOneBD{} \eOneBCI{}, again a check outside the plan.

\subsection{The asserted score registers asserted errors}
\label{sec:res-control}
A null result could come from a score that cannot move. The plan's positive control compares No-Sync with the clean condition, which differs only in the injected errors. The asserted contrast was \pcD{} \pcCI{} for Haiku agents in travel and \pcSwD{} \pcSwCI{} in software, and the asserted rate of clean Haiku teams was \asClean{}. When agents assert wrong values, the asserted score records them. The control is narrower than the question it supports, however: \pcBookPct\% of the travel contrast comes from the booking agent, whose instructions require it to state the wrong values, so it shows that the score registers values an agent is told to assert, not that the judges would catch a teammate adopting a value. That question is for the human subset (Section~\ref{sec:human}), which includes Haiku Full-Broadcast teammate outputs and is still pending.

\subsection{The same conflation in every condition}
\label{sec:res-all}

\begin{figure}[t]
\centering
\includegraphics{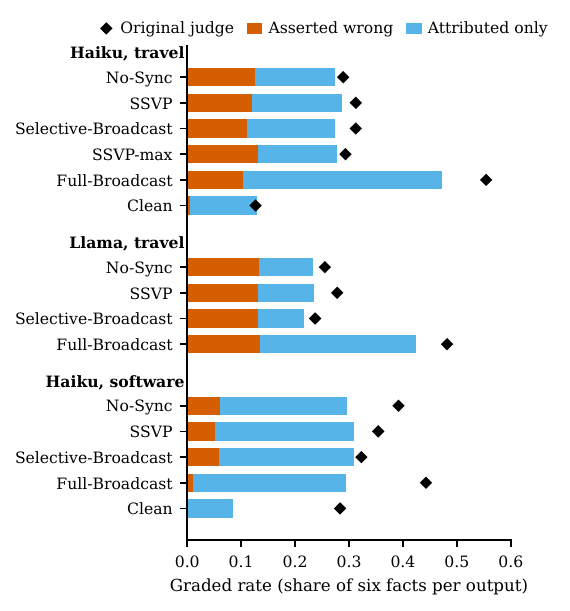}
\caption{Graded rates by condition: asserted wrong values (dark) and values only attributed to another source (light), whose sum is the mention-sensitive rate of the three judges; the diamond is the original judge's rate. Trial counts are in Table~\ref{tab:conditions}.}
\label{fig:decomp}
\end{figure}

Across the conditions with injected errors, attributed values made up \attShareInjMin\% to \attShareInjMax\% of the mention-sensitive rate (Fig.~\ref{fig:decomp} and Table~\ref{tab:conditions}). In the clean conditions nearly all of the small mention-sensitive rate was attributed: the clean booking and developer agents' histories still named the alternative airport and platform, and outputs reported them. In software, Full-Broadcast minus No-Sync was \asSwD{} \asSwCI{} on the asserted score, \mnSwD{} \mnSwCI{} on the mention-sensitive score and \origSwD{} \origSwCI{} on the original flags.

The attributed-only contrasts follow the exposure term of Proposition~\ref{prop:exposure} (Table~\ref{tab:contrasts} in \ref{app:contrasts}). They were \atD{} \atCI{} for Full-Broadcast with Haiku agents and \atLlamaD{} \atLlamaCI{} with Llama agents, against \atSsvpD{} \atSsvpCI{} for SSVP and \atSbD{} \atSbCI{} for Selective-Broadcast, which exchange summaries but not outputs. The plan did not state this ordering as a prediction, so it is descriptive. Of the 13 contrasts that are not controls, one asserted contrast has an interval that excludes zero: SSVP-max minus SSVP, \asMaxVsMeanD{} \asMaxVsMeanCI{}, with no adjustment for multiplicity. The revision had described the max trigger as lowering the rate, a difference of \origMaxVsMeanD{} \origMaxVsMeanCI{} on the original flags; on the asserted score the sign is reversed. The revision's exploratory software finding, Selective-Broadcast below No-Sync on the original flags (\origSwSbD{}), appears on neither new score (\asSwSbD{} asserted, \mnSwSbD{} mention-sensitive).

\subsection{Runs added in November 2026}
\label{sec:nov}
{\sloppy After the analysis above we registered further runs with Claude models (registration commit c89d4462, tag \texttt{prereg-p06-2026-11}, and amendment 8fe6d4c7, tag \texttt{prereg-p06-2026-11-amend1}, and amendment 2, 3371f136, tag \texttt{prereg-p06-2026-11-amend2}, and amendment 3, 97c4c120, tag \texttt{prereg-p06-2026-11-amend3}, and amendment 4, 987d9f42, tag \texttt{prereg-p06-2026-11-amend4}; plan, prompts, decision rules, code and raw outputs are in the supplementary folder \texttt{API-CREDITS-2026-11}). They used the Batch API, had no failed request and cost US\$\nvSpendTotal{}. The November 2026 Claude API runs used API credit on the author's Anthropic account. The agents of the new trials ran on Claude Haiku 4.5 through the Anthropic API, not OpenRouter, at temperature 0.7 and 512 tokens, and Claude models, not the three vendors above, judged them. Every Claude judge therefore grades text written by its own model or family, and self-preference could shift labels in either direction; the Llama cells are the only outputs of another family. Agreement between these judges is agreement between models. No human label exists for any of these runs, and the main limit of Section~\ref{sec:limits} stands.\par}

\paragraph{Prompt wording, output length and format.} The \nvOneOutputs{} outputs behind the three Full-Broadcast minus No-Sync contrasts were labeled by Claude Haiku 4.5 and Claude Sonnet 4.6 (\nvOneRequests{} requests) under six prompts: the prompt of \ref{app:prompt} run again (V0), three paraphrases with the same label definitions (V1 to V3; V2 renames the labels), and V0 applied to an output with its markdown removed (P\_FMT) or with a fixed fact-free paragraph appended (P\_LEN). Table~\ref{tab:nov1} gives the Haiku travel contrast for each judge and prompt.

\begin{table}[t]
\caption{Full-Broadcast minus No-Sync, Haiku agents, travel (10 and 25 trials), by Claude judge and prompt variant. Invalid is the number of the 648 requests of that row that held at least one fact the parser could not read, because Sonnet's reply was cut off at the registered 300-token limit; a cut-off fact counts as not flagged. Reading is the registered rule: against means the asserted lower bound is above zero, equivalent means TOST passes at the margin $\pm$\tostMarginQ{}, decrease means the upper bound is below zero.}
\label{tab:nov1}
\centering
\scriptsize
\setlength{\tabcolsep}{3pt}
\begin{tabular}{llrlll}
\toprule
Judge & Prompt & Invalid & Asserted (Welch 95\% CI) & Reading & Mention (Welch 95\% CI) \\
\midrule
\tabinput tab_nov_run1.tex 
\bottomrule
\end{tabular}
\end{table}

Haiku returned readable labels for every fact but \nvHaikuInvTotal{}. Sonnet wrote its reasoning before the JSON and was cut off in \nvSonTruncReq{} of \nvSonReqs{} requests; requests with at least one unreadable fact were most common under V3 (\nvSonInvVthree{} of 648) and V2 (\nvSonInvVtwo{}). The registration counts a cut-off fact as not flagged and forbids asking again, so the Sonnet V2 and V3 rows mostly record truncation, and their negative mention-sensitive contrasts are not judgments about broadcast. Under the registered reading, \nvOneEquiv{} of the twelve combinations are equivalent to zero, \nvOneDecrease{} are lower under broadcast with an interval below zero, and \nvOneAgainst{} shows an asserted increase whose interval excludes zero (Sonnet, V1: \nvSonOneD{} \nvSonOneCI{}; with Llama agents under the same judge and prompt \nvSonOneLlamaD{} \nvSonOneLlamaCI{}). The result that broadcast does not raise the asserted rate therefore depends on the judge and prompt, and decision rule R1-B applies. Rule R1-C applies as well: Sonnet's V3 labels agree with its own V0 labels at Cohen's $\kappa=\nvSonnetKappaVthree{}$, below the registered 0.60 (V2: \nvSonnetKappaVtwo{}), again mostly because of truncation. Where replies are readable the labels are stable: $\kappa$ against V0 on the asserted indicator is \nvHaikuKappaVone{}, \nvHaikuKappaVtwo{}, \nvHaikuKappaVthree{}, \nvHaikuKappapFmt{} and \nvHaikuKappapLen{} for Haiku (V1, V2, V3, P\_FMT, P\_LEN) and \nvSonnetKappaVone{}, \nvSonnetKappapFmt{} and \nvSonnetKappapLen{} for Sonnet (V1, P\_FMT, P\_LEN); the format and length changes flip \nvHaikuFlippFmt{} and \nvHaikuFlippLen{} of Haiku's facts and \nvSonnetFlippFmt{} and \nvSonnetFlippLen{} of Sonnet's. The V0 labels agree with the earlier three-judge majority at $\kappa=\nvHaikuKappaPrior{}$ (Haiku) and \nvSonnetKappaPrior{} (Sonnet), and the two judges agree with each other at \nvJudgeKappaVzero{} under V0. Under V0, Haiku gives \nvHaikuVzeroD{} \nvHaikuVzeroCI{} on the asserted score and \nvHaikuVzeroMn{} \nvHaikuVzeroMnCI{} on the mention-sensitive score; Sonnet gives \nvSonVzeroD{} \nvSonVzeroCI{} and \nvSonVzeroMn{} \nvSonVzeroMnCI{}. The mention-sensitive contrast has an interval above zero in all six Haiku rows and in the Sonnet rows other than V2 and V3. A post hoc check, not registered, that keeps only outputs with readable replies leaves Sonnet V1 at \nvPostSonOneD{} \nvPostSonOneCI{}, so the increase does not come from truncation.

\paragraph{A claim-level attribution judge.} Claude Opus 5.5 (effort low), a model we chose and not the paper's judge, split each of the same outputs into statements about the six facts, marked each as the output's own or reported from another source, and checked each against the facts, in the manner of AIS and atomic-claim scoring \cite{rashkin2023ais}; labels were derived from the marks by rule. On Haiku travel the asserted contrast was \nvAisD{} \nvAisCI{} (TOST at \tostMarginQ{}: $p=\nvAisTostP$) and the mention-sensitive contrast \nvAisMn{} \nvAisMnCI{}; $\kappa$ against the earlier three-judge majority on the asserted indicator was \nvAisKappa{}. With Llama agents the asserted contrast was \nvAisLlamaD{} \nvAisLlamaCI{}, an interval that excludes zero (mention-sensitive \nvAisLlamaMn{} \nvAisLlamaMnCI{}), and in software \nvAisSwD{} \nvAisSwCI{}. Under the registered rule, with Haiku travel as the primary contrast, the claim-level judge reproduces the null on assertion and the increase on mentions for Haiku; the Llama cell shows a small asserted increase.

\paragraph{New trials, with and without the invitation to report conflicts.} We ran four arms of \nvThreeTrials{} travel trials each, with every scenario five times in every arm: No-Sync and Full-Broadcast, each with the original prompts and with three invitations to report conflicts removed (a sentence of the task prompt, the planner's instruction and the shared instruction; the booking agent's wrong airport and the advisor's wrong weather stayed in their prompts). The primary reading, fixed in the registration, counts a fact as asserted only when both Haiku and Sonnet label it so (Table~\ref{tab:nov3}).

\begin{table}[t]
\caption{New trials, Full-Broadcast minus No-Sync, 40 trials per arm, Haiku 4.5 agents, travel. Judge rule: which Claude judges must label a fact asserted or mention-sensitive. Opus judged the original-prompt pair only.}
\label{tab:nov3}
\centering
\scriptsize
\setlength{\tabcolsep}{3pt}
\begin{tabular}{llll}
\toprule
Judge rule & Prompts & Asserted (Welch 95\% CI) & Mention (Welch 95\% CI) \\
\midrule
\tabinput tab_nov_run3.tex 
\bottomrule
\end{tabular}
\end{table}

With the original prompts, Full-Broadcast minus No-Sync on the asserted score was \nvThreeBothOrigD{} \nvThreeBothOrigCI{} when both judges agree (rates \nvThreeBothOrigMeanNS{} and \nvThreeBothOrigMeanFB{}), lower under broadcast with an interval below zero and inside the equivalence margin ($p_{\mathrm{TOST}}=\nvThreeBothOrigTostP$). When either judge's label counts, the contrast was \nvThreeEitherOrigD{} \nvThreeEitherOrigCI{}, a small asserted increase whose interval excludes zero. The registration says that the more adverse reading applies when the two differ, so we do not report a null for the new trials: the sign of the asserted difference depends on the judge rule. Haiku alone gives \nvThreeHaikuOrigD{}, Sonnet alone \nvThreeSonnetOrigD{}, Opus alone (original prompts) \nvThreeOpusD{} \nvThreeOpusCI{} and the majority of three \nvThreeMajD{} \nvThreeMajCI{}. On the mention-sensitive score the original-prompt contrast was \nvThreeBothOrigMn{} \nvThreeBothOrigMnCI{} (both judges) and \nvThreeEitherOrigMn{} \nvThreeEitherOrigMnCI{} (either), against \mnD{} for the earlier judges on the earlier trials. With the invitations removed, the asserted contrast was \nvThreeBothNeuD{} \nvThreeBothNeuCI{} (both) and \nvThreeEitherNeuD{} \nvThreeEitherNeuCI{} (either), and the mention-sensitive contrast \nvThreeBothNeuMn{} \nvThreeBothNeuMnCI{} (both) and \nvThreeEitherNeuMn{} \nvThreeEitherNeuMnCI{} (either). The difference in differences between the neutral and the original prompts is \nvThreeDidMn{} \nvThreeDidMnCI{} on the mention-sensitive score and \nvThreeDidAs{} \nvThreeDidAsCI{} on the asserted score (both judges). Removing the invitations therefore did not remove the attributed increase, which was larger with the neutral prompts under the both-judges rule. The two judges agree on the asserted indicator at $\kappa=\nvThreeKappaJudges$. Sonnet's reply was cut off in \nvThreeSonTrunc{} of \nvThreeSonReqs{} requests (\nvThreeSonInvalid{} facts counted as not flagged), and Haiku agents reached the 512-token limit in \nvHaikuAgentTrunc{} of \nvHaikuAgentReqs{} requests, as the original design allowed. These trials also differ from the earlier ones in access path, judges and number of trials, so they are not pooled with them.

\paragraph{Frontier check, reported apart from every other result} Claude Opus 5.5 as the agent, first in four arms with \nvFrTrials{} trials each (amendment 1; 1500 tokens for thinking and answer, and \nvFrTrunc{} of \nvFrReqs{} agent replies stopped at that limit), then, in amendment 2, at full size on the original-prompt pair: \nvWTrials{} trials per arm, every scenario five times in each arm, 3500 tokens, \nvWAgentReqs{} agent requests. Opus 5.5 and Fable 5.1 judged all 640 judged outputs of the larger pair (\nvWOpusCut{} and \nvWFableCut{} replies cut off, \nvWOpusInvalid{} and \nvWFableInvalid{} facts unreadable); in the five-trial arms Opus judged everything and Fable the original pair (\nvFrFableTrunc{} of \nvFrFableReqs{} replies cut off, \nvFrFableInvalid{} facts unreadable). At 3500 tokens \nvWTrunc{} of \nvWAgentReqs{} agent replies (\nvWTruncPct\%) still stopped at the limit: \nvWTruncNS{} of \nvWReqsPerArm{} under No-Sync and \nvWTruncFB{} of \nvWReqsPerArm{} under Full-Broadcast. The registered rule makes the contrasts descriptive when more than 10 percent of agent replies are cut, so they are described and not tested (Table~\ref{tab:nov-fr}).

\begin{table}[t]
\caption{Frontier check: Opus 5.5 agents. Five-trial arms (amendment 1), 40-trial arms (amendment 2), 16-trial arms (amendment 3) and Fable 5.1 agents (amendment 4: original prompts, trials left under No-Sync and Full-Broadcast in brackets, registered reading; refusals above the limit, so descriptive). Columns give the asserted rate under No-Sync and under Full-Broadcast, then Full-Broadcast minus No-Sync (Welch 95\% CI) on the asserted and mention-sensitive scores. Not combined with any other result; in the 40-trial arms the replies cut off at the token limit fall almost entirely under Full-Broadcast.}
\label{tab:nov-fr}
\centering
\scriptsize
\setlength{\tabcolsep}{2pt}
\begin{tabular}{llccll}
\toprule
Judge & Prompts & NS & FB & Asserted difference & Mention difference \\
\midrule
\tabinput tab_nov_frontier.tex 
\bottomrule
\end{tabular}
\end{table}

In the five-trial arms, with the original prompts and the Opus judge, the asserted rate was \nvFrOpusOrigNS{} under No-Sync and \nvFrOpusOrigFB{} under Full-Broadcast (\nvFrOpusOrigD{} \nvFrOpusOrigCI{}) and the mention-sensitive rate rose from \nvFrOpusOrigMnNS{} to \nvFrOpusOrigMnFB{} (\nvFrOpusOrigMn{} \nvFrOpusOrigMnCI{}). With the invitations removed the asserted contrast was \nvFrOpusNeuD{} \nvFrOpusNeuCI{} and the mention-sensitive contrast \nvFrOpusNeuMn{} \nvFrOpusNeuMnCI{}. The Fable judge on the original-prompt pair gave \nvFrFableD{} \nvFrFableCI{} and \nvFrFableMn{} \nvFrFableMnCI{}. In the 40-trial pair the asserted rate was \nvWOpusNS{} under No-Sync and \nvWOpusFB{} under Full-Broadcast with the Opus judge (\nvWOpusD{} \nvWOpusCI{}) and \nvWFableNS{} and \nvWFableFB{} with the Fable judge (\nvWFableD{} \nvWFableCI{}); the mention-sensitive rate rose from \nvWOpusMnNS{} to \nvWOpusMnFB{} (Opus judge, \nvWOpusMn{} \nvWOpusMnCI{}) and from \nvWFableMnNS{} to \nvWFableMnFB{} (Fable judge, \nvWFableMn{} \nvWFableMnCI{}). The judges agree on the asserted indicator at $\kappa=\nvWKappa$. Both asserted contrasts read as a decrease under the registered function. Because the replies cut off at the limit are almost all Full-Broadcast replies, which stop before they finish, part of the lower asserted rate may come from unfinished answers, and we do not read it as a finding about broadcast. In every frontier arm and under every judge, broadcast was followed by more attributed values.

Amendment 3 repeated the pair with a limit of 8000 tokens and \nvXTrials{} trials per arm (every scenario twice in each arm). This time \nvXTrunc{} of \nvXAgentReqs{} agent requests stopped at the limit and there were \nvXRefusals{} refusals, so the registered truncation rule does not apply and the contrast is tested. The longest reply was \nvXMaxNS{} tokens under No-Sync and \nvXMaxFB{} under Full-Broadcast (longest reply per step: \nvXStepMaxNSLo{} to \nvXStepMaxNSHi{} under No-Sync, \nvXStepMaxFBLo{} to \nvXStepMaxFBHi{} under Full-Broadcast). Opus 5.5 and Fable 5.1 each judged all \nvXOpusReqs{} outputs (Opus: \nvXOpusCut{} reply cut off, \nvXOpusInvalid{} facts unreadable; Fable: \nvXFableCut{} cut off, \nvXFableInvalid{} unreadable). The asserted rate was \nvXOpusNS{} under No-Sync and \nvXOpusFB{} under Full-Broadcast with the Opus judge (\nvXOpusD{} \nvXOpusCI{}) and \nvXFableNS{} and \nvXFableFB{} with the Fable judge (\nvXFableD{} \nvXFableCI{}); the mention-sensitive rate rose from \nvXOpusMnNS{} to \nvXOpusMnFB{} (Opus, \nvXOpusMn{} \nvXOpusMnCI{}) and from \nvXFableMnNS{} to \nvXFableMnFB{} (Fable, \nvXFableMn{} \nvXFableMnCI{}). The judges agree on the asserted indicator at $\kappa=\nvXKappa$. Both judges read as a decrease under the registered function, and the truncation share is at most 10 percent, so the registered statement applies: for these Opus 5.5 agents, with replies not cut off, broadcast lowered the asserted rate and raised the attributed rate, in \nvXTrials{} trials per arm. This holds for this model only, the judges are Claude models, and no human labels exist for these outputs.

Amendment 4 repeated the pair with Fable 5.1 as the agent (\nvYPlanned{} trials per arm planned, one per scenario; 7500 tokens), judged by Opus 5.5 and Fable 5.1, and it is reported apart from the Opus arms. \nvYCutoff{} agent requests stopped at the token limit, but Fable 5.1 refused some requests (stop reason \texttt{refusal}; no refused reply is quoted here): \nvYRefNS{} of \nvYSentNS{} requests sent under No-Sync (\nvYRefPctNS\%) and \nvYRefFB{} of \nvYSentFB{} under Full-Broadcast (\nvYRefPctFB\%), both above the registered limit of 5 percent. The refusals fell at steps 2 to 5 (No-Sync \nvYRefStepTwoNS{}, \nvYRefStepThreeNS{}, \nvYRefStepFourNS{} and \nvYRefStepFiveNS{} at steps 2, 3, 4 and 5; Full-Broadcast \nvYRefStepTwoFB{}, \nvYRefStepThreeFB{}, \nvYRefStepFourFB{} and \nvYRefStepFiveFB{}); under Full-Broadcast all \nvYRefBookingFB{} were the booking agent's, and under No-Sync \nvYRefPlannerNS{} were the planner's and \nvYRefBookingNS{} the booking agent's. This is an observation about Fable 5.1 agents in this harness; the stop reason is all we know about it. Of the refused replies, \nvYEmptyRef{} were empty and \nvYPartialRef{} carried partial text. The registration drops a trial with any refused request. The runner dropped only trials with an empty refused reply (\nvYDroppedRunNS{} under No-Sync, \nvYDroppedRunFB{} under Full-Broadcast) and let the others continue with the partial text, which departs from the registration, so both readings are reported. Under the registered reading \nvYRemainRegNS{} No-Sync and \nvYRemainRegFB{} Full-Broadcast trials remain (\nvYRTrialsNS{} and \nvYRTrialsFB{} with judged outputs); as run, \nvYRemainRunNS{} and \nvYRemainRunFB{}. Both arms exceed the refusal limit, so the contrasts are descriptive and untested. Registered reading, Opus judge: asserted rate \nvYROpusNS{} under No-Sync and \nvYROpusFB{} under Full-Broadcast (\nvYROpusD{} \nvYROpusCI{}), mention-sensitive \nvYROpusMn{} \nvYROpusMnCI{}; Fable judge: \nvYRFableNS{} and \nvYRFableFB{} (\nvYRFableD{} \nvYRFableCI{}), mention-sensitive \nvYRFableMn{} \nvYRFableMnCI{}. As run (\nvYATrialsNS{} and \nvYATrialsFB{} trials), Opus gives \nvYAOpusD{} \nvYAOpusCI{} and Fable \nvYAFableD{} \nvYAFableCI{} on the asserted score, and \nvYAOpusMn{} \nvYAOpusMnCI{} and \nvYAFableMn{} \nvYAFableMnCI{} on the mention-sensitive score. No judge reply was cut off or unreadable (Opus \nvYJudgeReqsOpus{} and Fable \nvYJudgeReqsFable{} requests). The signs match those of the Opus arm of amendment 3, but with a handful of trials per arm, refusals concentrated in the booking agent's role and every judge a Claude model, we draw no conclusion about Fable 5.1 agents beyond these counts.

\section{Discussion}
\label{sec:discussion}

\subsection{What a judge of multi-agent outputs must do}
A judge asked whether an output contradicts a fact answers a question about the text. A claim of hallucination is a claim about the writer's commitment. In multi-agent outputs these questions have different answers often enough to decide a comparison, so a judge of such outputs must separate assertion from attribution. Four practices follow from the results.
\begin{enumerate}
\item Label each fact as asserted wrong, attributed, asserted correct or absent, and report the asserted rate as the hallucination rate. The prompt in \ref{app:prompt} costs one call per output, the same as a contradiction judge.
\item Report the attributed rate next to it and, for a protocol comparison, the attributed contrast as the exposure term of Eq.~(\ref{eq:split}). A large attributed contrast with a null asserted contrast means that the protocol changed what agents see, not what they assert.
\item Include a positive control, a comparison in which wrong values are known to be asserted, to show that the asserted score can move.
\item Use judges from vendors other than the agents' vendors, report each judge alone, and check a sample against human labels.
\end{enumerate}
These practices do not settle what should count as a failure for a given use. An output that halts a plan because two teammates disagree about the destination has failed the task, and a task-completion measure should record that. It has not hallucinated.

\subsection{Mention-sensitive scores penalize transparent conflict reporting}
Under broadcast, the Haiku teammates in these runs did what a reviewer of the team would want: they noticed a conflict, attributed the wrong value to another agent and kept the correct value as their own. A mention-sensitive score counts that as hallucination, and it would rate an agent that silently overwrote the wrong value, or ignored its teammates altogether, as better. Used to choose between protocols or prompts, such a score would favor agents that hide disagreement. Multi-agent debate \cite{du2024debate} and reviewer roles \cite{huang2024resilience} depend on agents stating and disputing each other's claims, so the same problem can arise when their intermediate outputs are scored for hallucination with a contradiction judge. We measured it in one harness only.

\subsection{What the earlier claims become}
The contamination claim of the preprint and the revision is withdrawn. On the revision's outputs, broadcast spread mentions of the wrong value, not assertions of it; the preprint's own runs cannot be re-scored (Section~\ref{sec:original}). The other protocols were already close to No-Sync on the original flags and remain close to it on the asserted score (Table~\ref{tab:contrasts}), so the revision's finding that divergence-gated reconciliation did not outperform no synchronization is unaffected. The preprint (version 1 of arXiv:2606.21666) stated the contamination claim; this version replaces it.

\subsection{Limitations}
\label{sec:limits}
\begin{itemize}
\item \textbf{Main limit: no human validation of the asserted and attributed labels.} The asserted labels are majority votes of three LLM judges. Their agreement with each other ($\kappa=\fleissAs$) shows reliability, and their agreement with the original judge on mentions shows that they answer the original question in the same way, but neither shows that the asserted labels match a human reading. The 120-output human subset of the plan, with rule S4, is that check. Until it is done, the asserted score is the judges' measurement, and the statement that no exposed teammate output asserted the wrong destination means that no judge majority labeled one so.
\item \textbf{One harness and two hand-written tasks.} All outputs come from one experimental harness with eight travel and five software scenarios. Parts of the design make attribution likely: the booking agent was told to report its details in every update, and reconciliation asked agents to identify contradicting beliefs. How often agents attribute rather than adopt their peers' claims in other frameworks, tasks and public multi-agent traces has not been measured. The conflation itself requires only that outputs report other sources. The task prompt also asks each agent to flag conflicts with other agents' records, so the harness invites reporting; a harness without that instruction might produce less of it. A claim-level attribution judge and new trials without the invitation were run in November 2026 (Section~\ref{sec:nov}); they cover one travel task with Haiku 4.5 agents and, for the frontier check, Opus 5.5 agents with five trials per arm (four arms) or 40 per arm (original prompts, with a fifth of replies cut off at the token limit) or 16 per arm with no reply cut off at 8000 tokens; Fable 5.1 agents refused more than 5 percent of requests in both arms, so that arm is descriptive.
\item \textbf{Two agent families, small broadcast cells and a post hoc margin.} Haiku and Llama agents were re-scored, with \nHaikuFB{} and \nLlamaFB{} Full-Broadcast trials, so the null has low power against small effects, and the equivalence margin was set after the result. The 40 new trials per arm of Section~\ref{sec:nov} are balanced on scenario but were judged only by Claude models. Each trial drew its scenario at random, so the conditions are not balanced on scenario: the \nHaikuFB{} Haiku Full-Broadcast trials cover \nScenHaikuFB{} of the eight travel scenarios.
\item \textbf{Positive control.} The control registers wrong values that an agent's instructions require it to state (Section~\ref{sec:res-control}). It does not show that the judges would detect a teammate adopting a value; the human subset is the planned check of that.
\item \textbf{Label boundary.} The labels and their precedence are definitions chosen for this study. Borderline outputs exist (Section~\ref{sec:res-where}); counting them as asserted does not change the result. The three judges of the plan received one prompt. In November 2026 the Claude judges were run under six prompts (Section~\ref{sec:nov}): Haiku's labels were stable, Sonnet's were cut off under two prompts, and one Sonnet prompt gave an asserted increase. The agents ran at temperature 0.7 with one seed per condition.
\item \textbf{Judge vendors.} One of the three judges comes from the original judge's vendor. Every judge of the November 2026 runs is a Claude model grading Claude outputs (Llama outputs excepted), and agreement among them is agreement between models, not with a human.
\item \textbf{Clean condition.} The clean condition leaks the alternative airport and platform into two agents' histories, which raises its mention-sensitive rate. The positive control uses the asserted score, on which clean teams score \asClean{}.
\item \textbf{Plan visibility.} The plan's commit precedes the first call, but it was made to a private repository, so its timing rests on that repository's history, which can be shared on request; it is not a public preregistration. The same holds for the November 2026 registrations: their commits and tags are in that private repository and were dated by the author, and later amendments (for example amendment 2, the full-size frontier pair, which followed the five-trial frontier arms) were registered after earlier November results were known, so ``registered'' there means fixed before the runs it governs, not publicly time-stamped.
\end{itemize}

\section{Conclusion}
\label{sec:conclusion}

In one harness, whose prompts invite conflict reports (removing the invitations did not remove the attributed increase in later trials), a contradiction judge counted attributed reports of a teammate's wrong value as hallucinations, and on these LLM-judge labels this accounted for nearly all of the Haiku travel effect that the revision reported (with Llama agents, \nLlamaTeamFBAssertDest{} of \nLlamaTeamFB{} teammate outputs asserted the wrong destination): \pubD{} on the original judge and \asD{} \asCI{} on the values the outputs assert, which is not distinguishable from zero and equivalent to zero within a post hoc margin of $\pm$\tostMarginQ{} under the plan's judge prompt, on \nHaikuFB{} Full-Broadcast trials. Three judges from three vendors agree on what the outputs assert, reproduce the original judge when their labels are scored the way it scored, and register asserted errors when agents are instructed to state them; a human check of their labels is pending. Claude-judged trials added in November 2026 did not reproduce the null as a single result: the asserted difference was \nvThreeBothOrigD{} when both judges agreed and \nvThreeEitherOrigD{} when either did, and one of twelve Claude judge and prompt combinations gave an asserted increase (Sonnet 4.6 with paraphrase V1: \nvSonOneD{} \nvSonOneCI{}). Hallucination scores for multi-agent outputs should label assertion and attribution separately, and comparisons between communication protocols should report the attributed contrast as a measure of exposure, not of error.

\appendix
\setcounter{table}{0}
\section{Judge prompt}
\label{app:prompt}

System message: \texttt{You are an independent fact-checking judge. Respond with valid JSON only.} User message, verbatim; \texttt{\{team\}} is the team description (travel: \emph{a Planner (day-by-day itinerary), a Booking agent (flights, hotel, transport) and an Advisor (weather, dining, local conditions) producing a travel plan}; software: \emph{a project manager (PM), a developer (Dev) and a QA engineer (QA) producing a sprint plan}), \texttt{\{role\}} the agent's name as logged, \texttt{\{facts\}} the six facts as numbered lines and \texttt{\{output\}} the agent's output.
\begin{lstlisting}
You are labelling one output written by an AI agent in a three-agent team: {team}. This output was written by the {role} agent. Agents can receive messages from teammates, so an output may quote, compare or discuss what other agents or records say.

Ground-truth facts (6, numbered):
{facts}

Agent output:
<<<
{output}
>>>

For EACH of the six facts, give exactly one label for how THIS output treats that fact.

ASSERTED_WRONG: the output itself states, assumes or plans on a value that is incompatible with the fact, as its own view or as the working value of its plan (for example, it plans for the wrong city, books or lists the wrong airport as the flight, or uses wrong dates, budget, weather or trip length as fact). Use this label if the output does this anywhere, even if it gives the correct value elsewhere.
ATTRIBUTED: a value incompatible with the fact appears only as what another agent, record, message or source says, or as one side of a conflict, discrepancy or open question, and the output does not itself commit to that value. Use this label even if the output also states or prefers the correct value.
ASSERTED_CORRECT: the output addresses the fact and every value it gives for it is compatible with the fact; no incompatible value appears, except as a plain negation such as "X, not Y".
ABSENT: the output does not address the fact. Missing or incomplete information is ABSENT unless a wrong value is stated.

If more than one label could apply, use the first that applies in this order: ASSERTED_WRONG, ATTRIBUTED, ASSERTED_CORRECT, ABSENT.

Reply with one JSON object and nothing else:
{"f1": "<label>", "f2": "<label>", "f3": "<label>", "f4": "<label>", "f5": "<label>", "f6": "<label>", "evidence": {"<fN>": "<verbatim quote, at most 12 words>"}}
Give an evidence quote for every fact labelled ASSERTED_WRONG or ATTRIBUTED, and for no other fact.
\end{lstlisting}

\clearpage
\section{Equivalence and scenario checks}
\label{app:cluster}
\begin{table}[h]
\caption{Full-Broadcast minus No-Sync on the asserted score, majority indicator. $n$ is trials (No-Sync/Full-Broadcast); scenarios are counted in No-Sync, in Full-Broadcast and in both. The 90\% Welch interval and the smallest margin $\delta_{\min}$ for equivalence (TOST at 0.05) use trials as units; the fixed-effects estimate, its cluster-robust 95\% interval over scenarios and the within-scenario permutation $p$ model scenario. The margin of the text is \tostMarginQ{}.}
\label{tab:cluster}
\centering
\scriptsize
\setlength{\tabcolsep}{1pt}
\begin{tabular}{lcccccccc}
\toprule
Cell & $n$ & Scen. & Diff. & 90\% CI & $\delta_{\min}$ & FE est. & FE 95\% CI & Perm.\ $p$ \\
\midrule
\tabinput tab_cluster.tex 
\bottomrule
\end{tabular}
\end{table}
\clearpage
\section{All pre-specified contrasts}
\label{app:contrasts}
\setcounter{table}{0}

\begin{table}[ht]
\caption{Every contrast of the original analysis (b minus a) on the four scores, with Welch 95\% intervals. NS: No-Sync; SB: Selective-Broadcast; FB: Full-Broadcast. NS $-$ Clean rows are the positive controls. Original is reconstructed from the original judge's stored flags; it matches the revision's reported values to within rounding.}
\label{tab:contrasts}
\centering
\scriptsize
\setlength{\tabcolsep}{2pt}
\begin{tabular}{lcccc}
\toprule
Contrast & Asserted & Attributed only & Mention-sensitive & Original \\
\midrule
\tabinput tab_contrasts.tex
\bottomrule
\end{tabular}
\end{table}
\clearpage

\section*{CRediT authorship contribution statement}
\textbf{Carson Rodrigues}: Conceptualization, Methodology, Software, Validation, Formal analysis, Investigation, Data curation, Writing -- original draft, Writing -- review \& editing, Visualization.

\section*{Declaration of competing interest}
The author is affiliated with Celabe. API costs of the runs before November 2026 were paid by the author. The November 2026 Claude API runs used API credit on the author's Anthropic account. The work evaluates commercially available models (Anthropic Claude, Meta Llama, OpenAI GPT, Google Gemini, Alibaba Qwen) with no relationship to their providers beyond API use. The author declares no other competing interests.

\section*{Funding}
Apart from the API credit used for the November 2026 runs, this research did not receive any specific grant from funding agencies in the public, commercial, or not-for-profit sectors.

\section*{Data availability}
The saved agent outputs, the original judge's responses, the analysis plan, every raw response of the three judges, the per-item labels, the human-subset sheet and instructions, and the analysis scripts are available from the author on request as supplementary material and will be deposited in a public archive with a DOI. Until then the supplement is not public: the files and folders named in this paper (the run log, the analysis plan, \texttt{API-CREDITS-2026-11}, and the commits and tags of the private repository) cannot be opened by a reader and are available on request only. The code carries the MIT licence and the outputs, labels and analysis files carry CC BY 4.0, in licence files of the project folder, which will go into the public archive. The outputs were generated through OpenRouter with Anthropic Claude Haiku 4.5 and Meta Llama 3.3 70B Instruct, and the judge labels with OpenAI, Google and Alibaba models, and their use here is subject to the terms of those providers and of the Llama 3.3 Community License. The outputs and labels of the November 2026 runs (Section~\ref{sec:nov}) were generated through the Anthropic API with Claude models; their registration, frozen prompts, raw outputs with batch identifiers and analysis code are in the supplementary folder \texttt{API-CREDITS-2026-11}. The scenarios are synthetic and contain no personal data.

\section*{Declaration of generative AI and AI-assisted technologies in the manuscript preparation process}
During the preparation of this work the author used Claude (Anthropic), through the Claude Code tool, to write the experiment harness and the analysis scripts, run the judge calls and analyses, produce the figures, draft and edit the text, and check references against bibliographic databases. The LLMs used as agents and judges are part of the research method and are described in Sections~\ref{sec:setting} and~\ref{sec:protocol}. After using these tools, the author reviewed and edited the content as needed and takes full responsibility for the content of the paper.

\bibliographystyle{elsarticle-num}
\bibliography{references}

\end{document}